\pdfoutput=1

\documentclass[11pt]{article}
\usepackage{hyperref}

\usepackage[final]{acl}
\usepackage{tabularx}
\usepackage{times}
\usepackage{latexsym}
\usepackage{threeparttable}
\usepackage{enumitem}
\usepackage{longtable}
\usepackage{hyperref}

\usepackage[T1]{fontenc}

\usepackage[utf8]{inputenc}
\usepackage{pifont}
\usepackage{microtype}

\usepackage{inconsolata}

\usepackage{graphicx}
\usepackage{multirow}
\usepackage{multicol}
\usepackage{arydshln} 
\usepackage{booktabs}
\usepackage{colortbl}

\title{RealHiTBench: A Comprehensive Realistic Hierarchical Table Benchmark for Evaluating LLM-Based Table Analysis}

\author{
 \textbf{Pengzuo Wu\textsuperscript{1}\textsuperscript{*}},
 \textbf{Yuhang Yang\textsuperscript{1}\textsuperscript{*}},
 \textbf{Guangcheng Zhu\textsuperscript{1}},
 \textbf{Chao Ye\textsuperscript{1}},
 \textbf{Hong Gu\textsuperscript{2}},
\\
 \textbf{Xu Lu\textsuperscript{1}},
 \textbf{Ruixuan Xiao\textsuperscript{1}},
 \textbf{Bowen Bao \textsuperscript{1}},
 \textbf{Yijing He\textsuperscript{1}},
\\
 \textbf{Liangyu Zha\textsuperscript{3}},
 \textbf{Wentao Ye\textsuperscript{1}},
 \textbf{Junbo Zhao\textsuperscript{1}},
 \textbf{Haobo Wang\textsuperscript{1}\textsuperscript{\dag}}
\\
 \textsuperscript{1}Zhejiang University 
 \quad
 \textsuperscript{2}vivo Mobile Communication Co., Ltd
\\
 \textsuperscript{3}Institute of Computing Innovation, Zhejiang University
\\
 \texttt{\{wupengzuo, yangyuhang, wanghaobo\}@zju.edu.cn}
}

\begin{document}
\maketitle
\begin{abstract}
With the rapid advancement of Large Language Models (LLMs), there is an increasing need for challenging benchmarks to evaluate their capabilities in handling complex tabular data. However, existing benchmarks are either based on outdated data setups or focus solely on simple, flat table structures. In this paper, we introduce \textbf{RealHiTBench}, a comprehensive benchmark designed to evaluate the performance of both LLMs and Multimodal LLMs (MLLMs) across a variety of input formats for complex tabular data, including LaTeX, HTML, and PNG. RealHiTBench also includes a diverse collection of tables with intricate structures, spanning a wide range of task types. Our experimental results, using \textbf{25} state-of-the-art LLMs, demonstrate that RealHiTBench is indeed a challenging benchmark. Moreover, we also develop TreeThinker, a tree-based pipeline that organizes hierarchical headers into a tree structure for enhanced tabular reasoning, validating the importance of improving LLMs' perception of table hierarchies. We hope that our work will inspire further research on tabular data reasoning and the development of more robust models. The code and data are available at \url{https://github.com/cspzyy/RealHiTBench}. \footnote{\textsuperscript{*} Equal contribution.}\footnote{\textsuperscript{\dag} Corresponding author.}
\end{abstract}

\section{Introduction}

In recent years, there has been a lot of work on table analysis~\citep{table_reasoning_1, TAPERA}. Specifically, an increasing number of studies discuss table question answering (TableQA) tasks.~\citep{effective_distillation, table_sequence, table_recognition, Table_LLM_2, GraphOTTER}. Additionally, the introduction of large language models (LLMs), including close-source models such as GPT-4o~\citep{gpt} and Gemini-1.5-Pro~\citep{gemini}, and table-oriented models like TableGPT~\citep{tablegpt} further enhances models' capability of table comprehension. Given these advancements, the rapid development of LLMs and the continuous emergence of diverse tabular data in real-world scenarios have highlighted the need for a comprehensive evaluation of LLMs’ table understanding capabilities, putting forward new requirements on the relevant benchmarks.

    \begin{figure}
        \centering
        \includegraphics[width=1\linewidth]{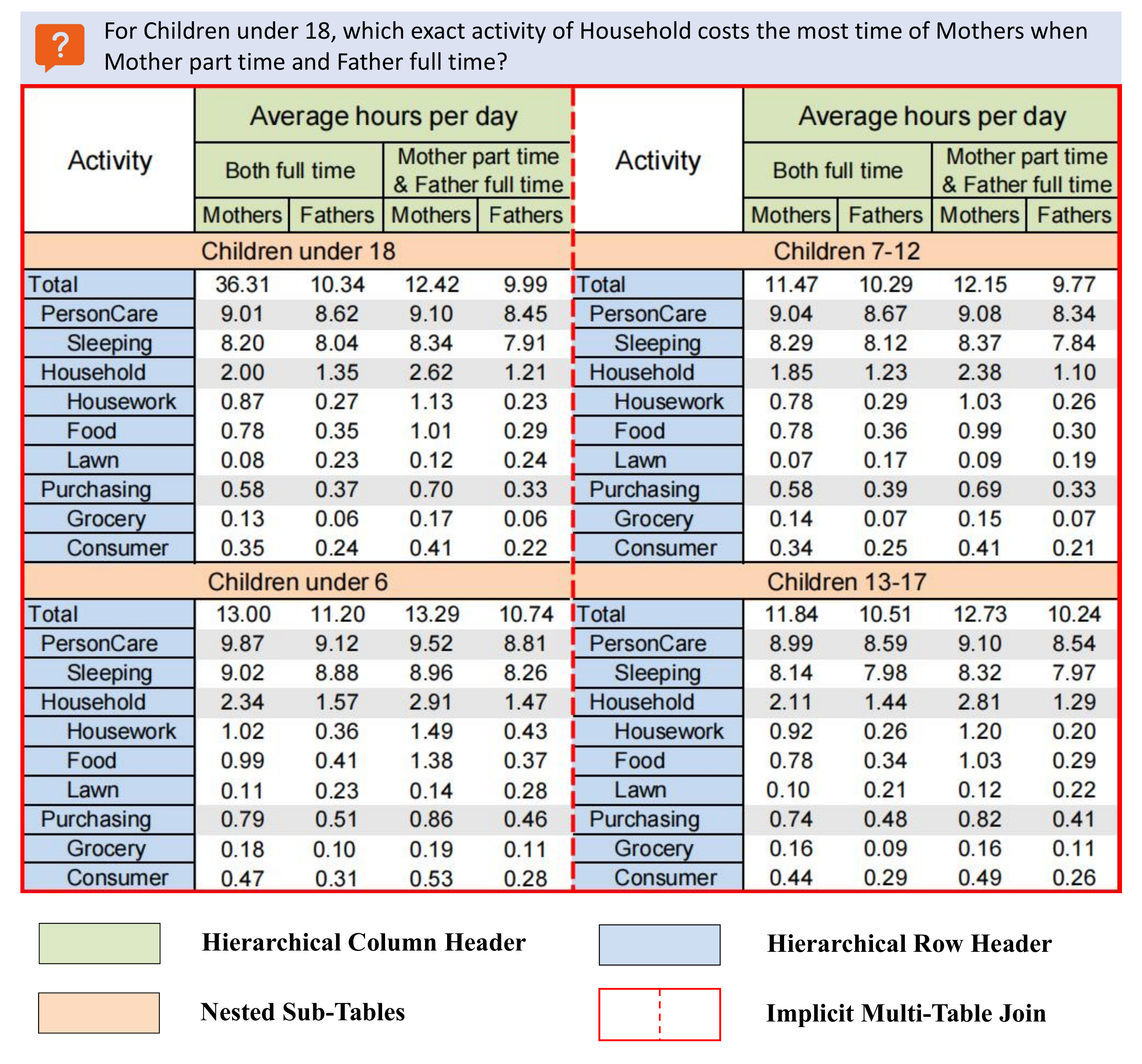}
        \caption{A complex table example with typical complex structures and a question-answer example.}
        \label{table_sample}
        \vskip -0.1in
    \end{figure}

Unfortunately, we find that the current tabular data benchmarks, such as TAT-QA~\citep{TAT-QA}, TableBench~\citep{tablebench}, and InfiAgent-DABench~\citep{infiAgent}, largely consist of flat tables. More concretely, in such tables, each column represents an attribute, each row represents a record, and all data are stored in a simple one-dimensional format; see Figure \ref{fig:tableBench_table_sample} for visual demonstration. However, in practical applications, humans often organize relatively complex tables to represent multifaceted relationships between variables. Such hierarchically structured tables are popular in various domains, including economy, science, and employment, on public data platforms.

To address this issue, some benchmarks have considered hierarchical tables, such as HiTab~\citep{HiTab} and SciTab~\citep{scitab}. However, these benchmarks fail to comprehensively present the LLMs' understanding capabilities of complex table structures. For example, SciTab~\citep{scitab} and AIT-QA~\citep{AIT-QA} only consider tables of the scientific and airline domains, respectively. MM-Tab~\citep{multimodal_table_understanding} provides only image-based input, while hierarchical tables in realistic applications are also presented in a textual way. SpreadSheetBench~\citep{SpreadsheetBench} focuses on the operation of tables. HiTab~\citep{HiTab} organizes its tables with lossy JSON format, only contains fundamental QA tasks, and provides incomplete supervision. More importantly, most of these benchmarks focus primarily on relatively simple hierarchical tables, particularly those with a basic column hierarchy, where the hierarchy typically does not exceed two levels. To date, there is still a lack of dedicated benchmarks for comprehensively evaluating LLMs' ability to understand complex tabular hierarchies.

To this end, we propose \textbf{RealHiTBench}, a challenging \textbf{Real}istic \textbf{Hi}erarchical \textbf{T}able \textbf{Bench}mark based on complex tables and tasks. \textbf{(i) Complex Table Structures:} Our benchmark includes complex tables with intricate features (partially depicted in Figure~\ref{table_sample}), which are commonly found in real-world scenarios but often overlooked in existing benchmarks. \textbf{(ii) Modal and Format Diversity:} We explore the performance of both text and image approaches while others focusing solely on one of them. Our benchmark provides LLMs and MLLMs to be tested with various input formats, such as LaTeX, HTML, and PNG. \textbf{(iii) Question Diversity:} RealHiTBench covers a wide range of question types, each designed to test different aspects of a model's ability. Especially a type called \textit{Structure Comprehending} is designed on the basis of complex parts. \textbf{(iv) Accurate and Efficient Annotation:} Our benchmark employs a rigorous annotation process. GPT-based automated annotation and human checks ensure the accuracy and reliability of the questions and answers.

We conducted a comprehensive evaluation of abundant types of models consisting of table-oriented LLMs, and open-source/closed-source generic LLM/MLLMs. In addition, we tested a couple of table reasoning tasks with different proper metrics. The performance of different models differs greatly, with average scores of all tasks ranging from 7.27 to 56.95. Importantly, overall low scores (mainly below 70) highlight that the ability of LLMs to comprehend and process intricate table structures remains an area in need of significant improvement. We also develop a tree-based pipeline (dubbed TreeThinker) that automatically injects table hierarchies into instructions. Empirically, we show that, \textit{with self-emphasized table hierarchies, LLMs' structural understanding ability can indeed be enhanced}.

\section{Related Work}

\paragraph{Table Analysis.} Table analysis is pivotal across numerous domains~\citep{text2analysis}, while table question answering (TableQA) can effectively assess analytical capabilities~\citep{origin_tableQA}. Recently, there has been a growing number of studies focusing on the ability of LLMs to comprehend tabular data~\citep{Table_LLM_1, Table_LLM_2, Table_LLM_3}. However, the development of LLMs leads to the emergence of different branches. Common tabular data are in text form to be processed by LLMs~\citep{gemini, qwen2.5, llama3}, while other image-based tables are suited to be processed by MLLMs~\citep{Qwen2-VL, llava}. Additionally, some table-oriented LLMs for both modalities have also been proposed~\citep{tablegpt, multimodal_table_understanding}. Therefore, the flourishing development of LLMs makes challenging benchmarks for tabular data increasingly important.

\paragraph{Table Analysis Benchmarks.} An increasing number of benchmarks for table analysis are discussed~\citep{TabFact, fetaQA}. However, a proportion of benchmarks focus more on other aspects, such as reasoning methods and supervision fine-tuning, while overlooking the structural complexity of the data, especially the table dimensions~\citep{tablebench, multimodal_table_understanding}. Although some recent benchmarks have introduced the concepts of hierarchical tables~\citep{HiTab, AIT-QA, SpreadsheetBench}, and a complex question answering benchmark has been proposed to bridge the gap between theoretical and real-world tabular data~\citep{tablebench}, the tables in these benchmarks are not complex enough and there has been no detailed discussion of the tables.

Therefore, in order to introduce a benchmark that can effectively evaluate current LLMs, we propose RealHiTBench consisting of complex tables and TableQA, which evaluates LLMs and MLLMs, as well as textual and visual input data. Furthermore, we look forward to the possible future development directions of LLMs in table reasoning.

\begin{table*}[!ht]
\centering
\caption{Comparison with existing datasets in Dataset Information, Task Types, and Input Formats. Here are abbreviations inside and their meanings: FC stands for Fact Checking, NR for Numerical Reasoning, DA for Data Analysis, CG for Chart Generation, and SC for Structure Comprehending. 
}
\renewcommand{\arraystretch}{1.3}
\label{tab:Main_Comparison}
\resizebox{\textwidth}{!}{
    \begin{tabular}{cccccccccccccc}
    \toprule[0.5mm]
    \multirow{2}{*}{Benchmark} & \multirow{2}{*}{Year} & \multicolumn{5}{c}{Dataset Information} & \multicolumn{5}{c}{Task Type}  & \multicolumn{2}{c}{Data Format} \\ 
    \cmidrule(lr){3-7} \cmidrule(lr){8-12} \cmidrule(lr){13-14}
     & & Hierarchy & $\mathbf{\#}$Table &  $\mathbf{\#}$QA pairs & Source & Domain & FC & NR & DA & CG & SC & Text & Image   \\
    \hline
    HiTab & 2022 &  High & 3,597 & 10,672 & 2 & 28 &  \ding{51} &  \ding{51} & \ding{56} & \ding{56} & \ding{56} &  \ding{51} & \ding{56} \\
    AIT-QA & 2022 &  High & 116 & 515 & 1 & 1 & \ding{56} &  \ding{51} & \ding{56} & \ding{56} & \ding{56} &  \ding{51} & \ding{56}\\
    MultiHierTT & 2022 &  High & 1,800 & 10,440 & 1 & 1 & \ding{56} &  \ding{51} & \ding{56} & \ding{56} & \ding{56} &  \ding{51} & \ding{56}\\
    \midrule
    TableBench & 2024 & Low & 586 & 3,544 & 6 & 18 &  \ding{51} &  \ding{51} &  \ding{51} &  \ding{51} & \ding{56} &  \ding{51} & \ding{56} \\
    InfiAgent-QABench & 2024 & Low & 52 & 257 & 1 & 9 & \ding{56} &  \ding{51} &  \ding{51} & \ding{56} & \ding{56} &  \ding{51} & \ding{56} \\
    TableVQA-Bench & 2024 & Low & 894 & 1,500 & 3 & 4 &  \ding{51} & \ding{51} & \ding{56} & \ding{56} & \ding{56} & \ding{51} & \ding{51} \\
    Text2Analysis & 2024 & Low & 347 & 2,249 & 1 & 1 & \ding{56} & \ding{56} &  \ding{51} &  \ding{51} & \ding{56} &  \ding{51} & \ding{56} \\
    \midrule
    \textbf{RealHiTBench} & \textbf{2025} &  \textbf{High} & \textbf{708} & \textbf{3,752} & \textbf{13} & \textbf{24} &  \ding{51} &  \ding{51} &  \ding{51} &  \ding{51} &  \ding{51} &  \ding{51} &  \ding{51} \\
    \bottomrule[0.5mm]
    \end{tabular}}
\end{table*}

\section{RealHiTBench}
\label{sec:RealHiTBench}
In practical applications, tabular data holds significant value. Numerous previous studies have proposed a series of benchmarks ~\citep{TAT-QA, ToTTo}, which have effectively propelled research in this field. However, as application scenarios become increasingly complex and large language models (LLMs) gradually emerge as the mainstream method for table reasoning~\citep{effective_distillation}, the previous benchmarks struggle to provide a comprehensive and accurate assessment of model capabilities. Hence, there is an urgent need for a new, realistic, and challenging benchmark to evaluate the progress of research methods in this field.
We propose RealHiTBench focusing on complex structures and tasks. Complex tables, including hierarchical tables, are widely used in various domains~\citep{scitab, AIT-QA}. We propose such a challenging benchmark full of complex tables and corresponding difficult tasks to assess the upper bound of the comprehension capability of LLMs, which may further inspire future research on processing tables with LLM. 

\subsection{Table Collection and Process}
To construct a realistic and comprehensive dataset, all tables in our benchmark are raw from 13 different open platforms and cover 24 different domains like economy, society, science, and so on (shown in the Appendix~\ref{appendix:sources} and \ref{appendix:domains}). 
In real applications, table-related benchmarks are presented in either textual ~\citep{HiTab, ToTTo} or visual ~\citep{TableVQA-Bench} manner. However, there is still no fixed format for complex table understanding. Therefore, we explore the influence of textual and visual ways. Specifically, we also explore the influence of different input formats, including LaTeX, HTML, CSV, Markdown, and PNG, whose performance is shown in Figure~\ref{appendix:formats}.

\begin{figure}
    \centering
    \includegraphics[width=1\linewidth]{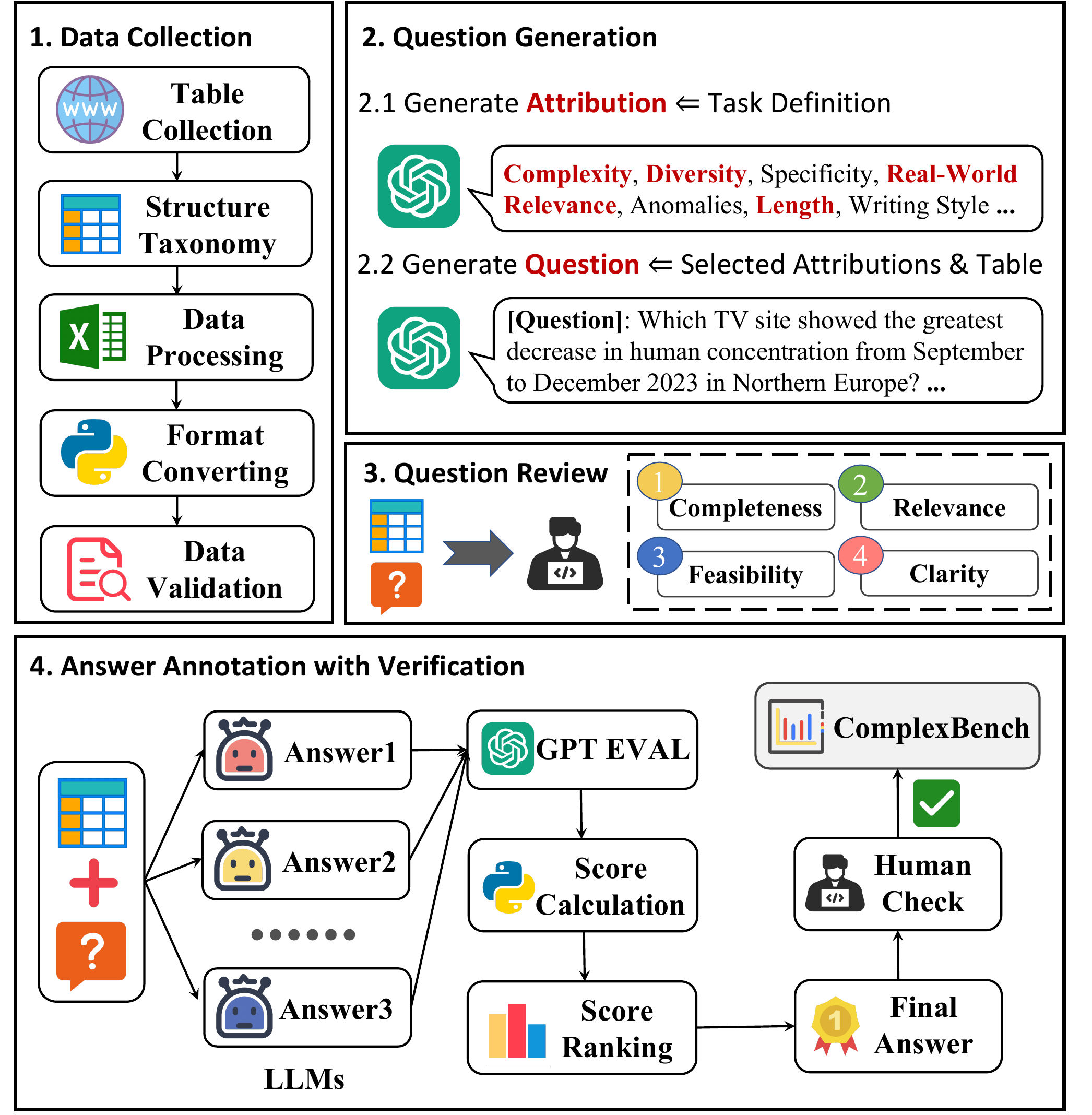}
    \caption{The construction process of \textbf{RealHiTBench}. }
    \vskip -0.1in
\end{figure}

\subsection{Complex Structures} In actual scenarios, we often encounter tables that show structural and even semantic complexity. However, current benchmarks \citep{HiTab, scitab} have not adequately discussed the complexity of table structure. Concentrating on complexity in tabular data, we define 5 complex structure types as following definitions. Most types of complex structure are presented in Figure~\ref{table_sample}.

\paragraph{(1) Hierarchical Column Header.} A column header, typically located at the top of a table, serves as the title for each column. It identifies the category, attribute, or subject of the data in that column and provides a structured organization of the table data. In most cases, the feature of the complex column header is cell merging, which leads to the complex header hierarchy in the table (the cells with green background in Figure~\ref{table_sample}).

\paragraph{(2) Hierarchical Row Header.} A row header is a label or identifier for each row in a table, typically located at the beginning of the row. In realistic tables. Row headers sometimes use indentation or clustering to present classified semantics. The common hierarchical row header is presented by indentation within a single column (the cell groups in blue within Figure~\ref{table_sample}). Moreover, the hierarchical row header is also presented by multiple columns, with a large merged cell corresponding to several subcategory cells in horizontal direction.

\paragraph{(3) Nested Sub-Tables.} Sometimes, due to semantic requirements, a whole table consists of several areas with vertical segmentations. Typical presentation is that there are horizontal cells spanning the full width of the table (the cells in orange within Figure~\ref{table_sample}). These full-width cells divide the root table into nested sub-tables.

\paragraph{(4) Multi-Table Join.} Speaking of the dimensions of a table, single-table scenarios cannot necessarily reflect the complexity of real-world applications, and multi-table tasks have been proposed in recent work~\citep{MMQA}. We extend the representation of multiple tables to two refined categories: \textit{(i) Explicit Multi-Table Join:} We define the majority of the multi-tables that have been discussed as Explicit Multi-Table (shown in Figure~\ref{fig:MMQA_multi_table_sample}). \textit{(ii) Implicit Multi-Table Join:} However, we notice other special multi-tables in our dataset. As depicted with red box in Figure~\ref{table_sample}, there are sub-tables with the same structures, especially column headers. The implicit Multi-Table type looks not different from a normal single table, but in fact, it is in the form of multiple tables. Interestingly, this type sometimes contains additional semantics, such as comparisons, but is not easy to detect when comprehending tables. 

\paragraph{(5) Miscellanies.} We also find that some other special contents obtain a part of the tabular information, including \textit{additional explanatory texts} outside the table and \textit{cell background colors}. These nonstructural elements also play a certain role in complex tables.

\paragraph{Remark.} It is worth noting that the hierarchical information we consider also appears in some existing benchmarks. However, we include tables with higher complexity (see Table~\ref{tab:complexity_comparison} for more details). Notably, it is necessary to clarify some differences between our work and the previous HiTab~\citep{HiTab} benchmark. First, HiTab pre-extracts the tree structure of hierarchical tables in JSON format, which prevents effective evaluation of whether LLMs can directly understand structural information from table inputs. Additionally, it suffers from a restricted focus on three domains, a simplistic QA task, and incomplete supervision.

\subsection{Complex Tasks}

In order to evaluate the model's abilities comprehensively, we define 5 primary types following TableBench~\citep{tablebench}: \textit{Fact Checking (FC)}, \textit{Numerical Reasoning (NR)}, \textit{Data Analysis (DA)}, \textit{Chart Generation (CG)}, and \textit{Structure Comprehending (SC)} (in the Table~\ref{tab:model_performance}). More detailed subtypes (shown in Appendix~\ref{appendix:task_types}) are derived from the above types. Notably, we extend one type named \textbf{\textit{Structure Comprehending}}, which is tailored for complex tables. This kind of task provides the new table by exchanging some complex parts of the source table. Then we ask LLMs the same question with two similar tabular input to evaluate the ability of LLMs to comprehend structures.

Based on the question types above, we meticulously design pretty difficult questions for each complex structure. We take questions generated from the table in Figure~\ref{table_sample} as an example: \textbf{(i) Header Hierarchy:} For the characteristics of header hierarchy, a possible question is: \textit{For Children under 18, which exact activity of Household cost the most time of Mothers when both full time?} It is difficult to identify the scale of columns and rows with classification and hierarchy. \textbf{(ii) Nested Sub-Tables:} several segments of a whole table lead to interesting semantic tricks. One challenging question is: \textit{How much total time costs Mothers in Both full time for all the children recorded in the table?} The correct conclusion is to just sum up total time for children under 18, instead of the whole table. Such summing questions with content partition tests if models thoroughly understand the inclusion relationships among sub-tables. 

\subsection{Annotation Generation}

\paragraph{Question Generation.} Following the previous work~\citep{Attributes}, we select some attributes that we also agree are important as the core for question construction from what GPT-4o considers. After feeding different prompts of each question type and acquiring generated questions, we conduct manual question review according to four factors: \textit{Relevance}, \textit{Completeness}, \textit{Feasibility}, and \textit{Clarity}. After the first turn of annotation, we swap questions of each other to check the rationality.

\paragraph{Answer Generation.} To eliminate the instability of GPT generation, we generate answers with GPT-4o based on well-designed answer prompt and innovate the pipeline: We feed a table with prompts to several models, acquiring multiple results and the frequency of occurrence of each result. Then the results are input into the \textit{G-Eval}~\citep{G-EVAL} module and scored for each result. Thus, final scores of each result can be figured out with initial score and frequency. The result with the highest final score becomes the candidate answer. We also conduct careful human check again that the table expert is dedicated to check answers one by one according to the question and exact table content. Furthermore, we employ a tree-based method to enhance our annotation process, finding that the automatic annotation accuracy is further improved.

\subsection{Dataset Statistics}

We propose RealHiTBench containing 708 tables and 3,752 questions. RealHiTBench completely focuses on discussing complex tables with such characteristics: \textbf{(1)} Tables in our benchmark are totally raw from available public platforms, and our benchmark totally consists of complex tables in structure or semantics. Thus, we organize 6 university students as annotators spending 150 hours each person on collecting these tables. By the way, annotators spend 480 hours each person on question answering annotations. \textbf{(2)} Compared to work involving complex tables such as HiTab~\citep{HiTab} and MultiHierTT~\citep{MULTIHIERTT} before, the tables in our dataset are more complex (quantized in Table~\ref{tab:complexity_comparison}), even the median sizes of the tables are larger than tables in other works ~\citep{tablebench}. \textbf{(3)} We systematically define different types of complex tabular structure and classify them for further exploration. \textbf{(4)} We introduce a couple of challenging task types. The tasks of \textit{structure comprehending} prove to be difficult for LLMs (shown in section~\ref{Main_Results}). The above characteristics are shown in Figure~\ref{table_sample} and Table~\ref{tab:Main_Comparison}.

\section{TreeThinker}
\label{TreeThinker}

\begin{figure}
    \centering
    \includegraphics[width=1\linewidth]{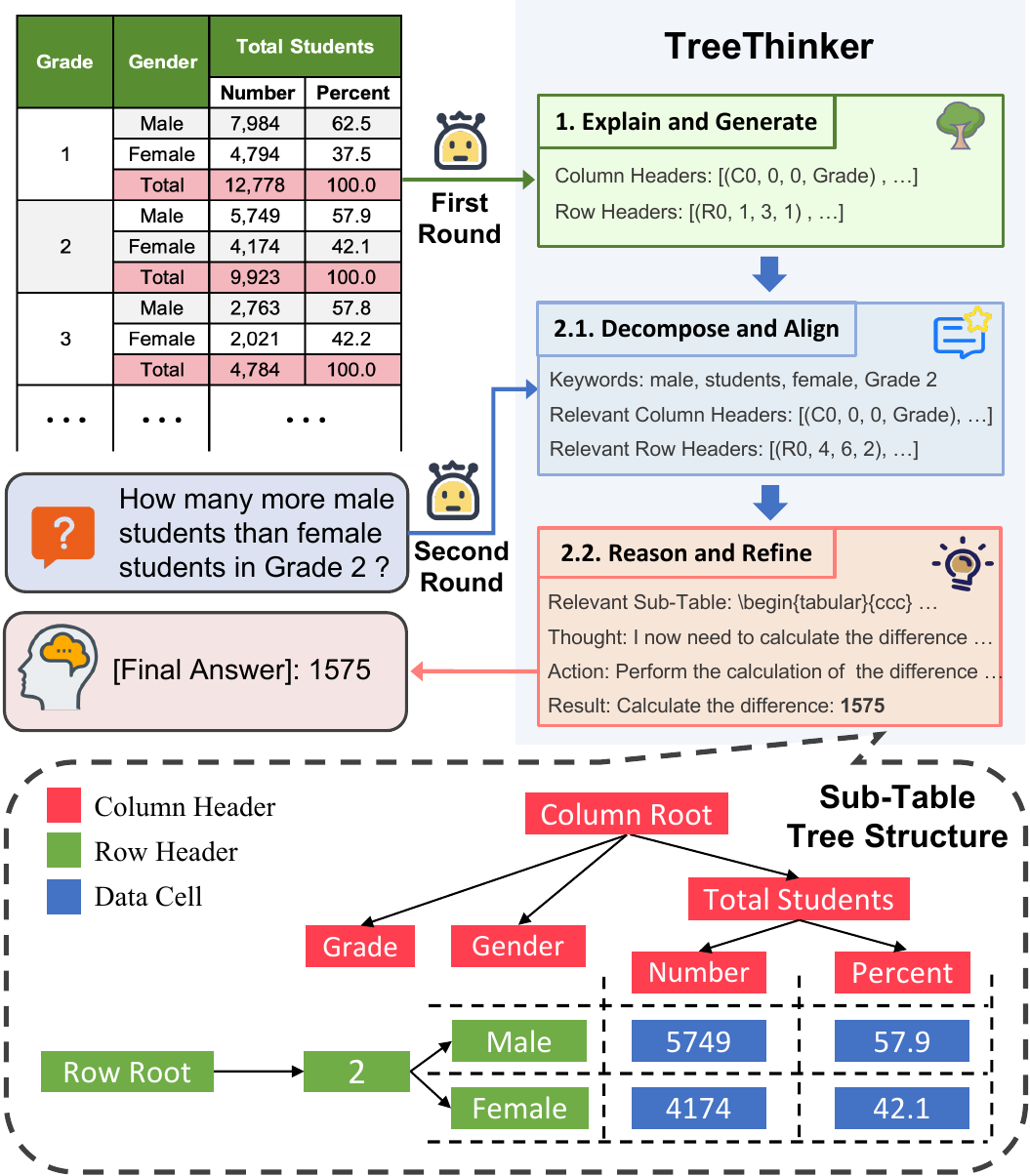}
    \caption{The overall framework of \textbf{TreeThinker}.} 
    \label{treethinker}
\end{figure}

Beyond benchmark, we propose \textbf{TreeThinker}, a understanding-augmented pipeline that enhances the model’s ability in complex hierarchical tables question answering. As shown in Figure \ref{treethinker}, we first prompt the model to \textbf{automatically} organizing hierarchical headers into \textbf{Tree Structure}, then aligning keywords from questions with the tree headers, and efficiently locating sub-table relevant to the question.

\subsection{Tree Generation}
Complex tables often contain multi-level row and column headers, making it difficult for models to accurately analyze their intricate hierarchical relationships. Therefore, to enhance the model’s ability in understanding complex table structures, we prompt model to explain the table’s headers structure and organize them into tree structure.

\paragraph{Explanation.} We first enable the model to self-explain the structural information of the table headers, including their meanings, scopes of influence, hierarchical structure, and relationships between headers. Referring to the previous approach \citep{TableTree}, we prompt the model by encoding each node of the table header as a tuples $T=(t_1,t_2,t_3,t_4)$. Specifically, the first element represents row header (R) or column header (C), along with its corresponding level. And the second and third elements represent its start and end positions, while the fourth element contains the content of the cell. 

For example, a tuple (R0, 1, 2, City) indicates that it is a row header (R) at level 0, spanning from row 1 to row 2, with the value City. This approach compresses header information into a tuple list $L = [T_1,T_2,\cdots,T_n]$, enables the model to more clearly identify the hierarchical structure and generating a "Structural Blueprint" of the table that effectively guides subsequent reasoning.

\paragraph{Generation.}
 The tree structure, with its clear parent-child connections, can accurately represents the hierarchical relationships of table headers, while making table data organization and retrieval more intuitive and efficient. Therefore, we ask the model to organize the scattered header list $L$ based on their hierarchical relationships into the Table-Header Tree $H$ with following steps: \textbf{(1)} Divide the tuple list $L$ into groups according to their levels, such that all tuples with the same level are grouped together. Add a special ROOT node Leveled "-1" for rows and columns. \textbf{(2)} For each tuple $A\in L$. If the start and end positions of \textit{A} are equal ($A_{t_2} = A_{t_3}$), mark \textit{A} as a leaf node. \textbf{(3)} Otherwise, compare its $t_2$ and $t_3$ values with every closest higher level and same flag tuple \textit{B}. If tuple \textit{A} is within the range of tuple \textit{B} ($A_{t_2}\geq B_{t_2}$ and $A_{t_3}\leq B_{t_3}$), then \textit{B} is the parent-header of \textit{A}. \textbf{(4)} Repeat steps 2 and 3 iteratively until all tuples in $L$ are linked to their respective parent nodes (Tuples without parent node are linked to the ROOT node), forming a hierarchical Table-Header Tree $H$.

\begin{table*}[!ht]
    \centering
    \caption{The evaluation results of advanced models with Text and Image Inputs on RealHiTBench.}
    \renewcommand{\arraystretch}{1.2} 
    \setlength{\tabcolsep}{4pt} 
    \resizebox{\textwidth}{!}{
    \begin{tabular}{cc>{\centering\arraybackslash}p{1.85cm}>{\centering\arraybackslash}p{1.85cm}>{\centering\arraybackslash}p{1.85cm}>{\centering\arraybackslash}p{1.85cm}>{\centering\arraybackslash}p{1.85cm}>{\centering\arraybackslash}p{1.85cm}>{\centering\arraybackslash}p{1.85cm}>{\centering\arraybackslash}p{1.85cm}>{\centering\arraybackslash}p{1.85cm}>{\centering\arraybackslash}p{1.85cm}}
        \toprule
        \multirow{2}{*}{\textbf{Model}} & \multirow{2}{*}{\textbf{Input}} & \multicolumn{2}{c}{\textbf{Fact Checking}} & \multicolumn{2}{c}{\textbf{Numerical Reasoning}} & \multicolumn{2}{c}{\textbf{Structure Comprehending}} & \multicolumn{2}{c}{\textbf{Data Analysis}} & \multicolumn{2}{c}{\textbf{Chart Generation}} \\
        \cmidrule(lr){3-12}
        \cmidrule(lr){3-4} \cmidrule(lr){5-6} \cmidrule(lr){7-8} \cmidrule(lr){9-10} \cmidrule(lr){11-12}
          &  & EM & F1 & EM & F1 & EM & F1 & GPT-EVAL & ROUGE & PASS@1 & ECR \\
        \midrule
        \rowcolor{gray!30} \multicolumn{12}{c}{\textbf{Table-oriented Models}} \\
        \textbf{TableGPT2-7B} & Text & \textbf{46.10} & \textbf{53.80} & \textbf{29.31} & \textbf{39.81} & \underline{48.23} & \underline{56.68} & \textbf{62.76} & \textbf{33.25} & \textbf{32.47} & \textbf{67.53} \\
        \textbf{TableLlama} & Text & 14.30 & 19.35 & 7.26 & 12.63 & 36.31 & 42.91 & 24.73 & 7.89 & 0 & 0 \\
        \textbf{TableLLM-Llama3.1-8B} & Text & 33.44 & 39.49 & 13.36 & 21.02 & \textbf{53.28} & \textbf{58.72} & \underline{47.86} & \underline{27.26} & \underline{6.49} & \underline{22.73} \\
        \textbf{TableLLM-Qwen2-7B} & Text & \underline{33.53} & \underline{40.41} & \underline{22.05} & \underline{29.04} & 36.09 & 44.75 & 47.30 & 18.86 & 4.55 & 10.39 \\
        \textbf{Table-LLava-7B} & Image & 4.19 & 7.05 & 2.98 & 5.59 & 7.38 & 11.83 & 22.46 & 8.72 & 0 & 1.30 \\
        \midrule
        \rowcolor{gray!30} \multicolumn{12}{c}{\textbf{Open-source Large Language Models}} \\
        \textbf{Llama3.1-8B-Instruct} & Text & 30.32 & 44.93 & 14.53 & 27.21 & 35.90 & 50.80 & \underline{60.12} & \underline{32.25} & 4.55 & 13.64 \\
        \textbf{Qwen2.5-7B-Instruct} & Text & 18.65 & 38.39 & 5.32 & 19.75 & 23.48 & 44.81 & 40.17 & 24.17 & \underline{15.58} & \underline{48.70} \\
        \textbf{Mistral-7B-Instruct-v0.3} & Text & 15.61 & 31.88 & 3.37 & 16.10 & 21.21 & 49.62 & 57.74 & 19.84 & 7.14 & 18.18 \\
        \textbf{Llama3.3-70B-Instruct} & Text & \textbf{53.08} & \textbf{64.53} & \textbf{36.58} & \textbf{48.99} & \textbf{55.81} & \textbf{68.93} & 52.26 & 27.98 & \textbf{24.03} & \textbf{50.65} \\
        \textbf{Qwen2.5-72B-Instruct} & Text & \underline{51.93} & \underline{62.15} & \underline{26.98} & \underline{39.23} & \underline{54.55} & \underline{68.34} & \textbf{68.45} & \textbf{35.90} & 14.29 & 27.27 \\
        \midrule
        \rowcolor{gray!30} \multicolumn{12}{c}{\textbf{Open-source Multimodal Large Language Models}} \\
        \textbf{LLaVa-v1.5-7B} & Image & 5.51 & 9.37 & 1.17 & 6.54 & 13.23 & 30.51 & 26.60 & 17.74 & 0 & 11.04 \\
        \textbf{mPLUG-Owl3-7B} & Image & 8.71 & 14.34 & 4.02 & 12.72 & 41.24 & 48.86 & 30.71 & 18.53 & 3.25 & 6.49 \\
        \textbf{Llama3.2-11B-Vision-Instruct} & Text & 18.41 & 38.44 & 9.86 & 22.97 & 23.39 & 43.22 & 52.75 & 30.59 & 4.55 & 17.53 \\
        \textbf{Llama3.2-11B-Vision-Instruct} & Image & 15.84 & 26.55 & 9.59 & 18.49 & 20.87 & 32.57 & 41.32 & 22.75 & 1.95 & 20.78 \\
        \textbf{Llama3.2-11B-Vision-Instruct} & Image+Text & 19.39 & 32.54 & 10.89 & 20.94 & 27.99 & 43.85 & 39.98 & 22.43 & 6.49 & \underline{35.71} \\
        \textbf{Llama3.2-90B-Vision-Instruct} & Text & \underline{52.10} & \underline{61.84} & \underline{28.15} & \underline{41.21} & \underline{58.52} & \underline{69.57} & \textbf{57.28} & \underline{31.87} & \underline{7.79} & 22.73 \\
        \textbf{Llama3.2-90B-Vision-Instruct} & Image & 23.75 & 36.08 & 25.46 & 28.05 & 33.33 & 46.40 & 41.18 & 25.46 & 5.19 & 16.23 \\
        \textbf{Llama3.2-90B-Vision-Instruct} & Image+Text & \textbf{55.60} & \textbf{65.19} & \textbf{38.65} & \textbf{49.71} & \textbf{59.80} & \textbf{70.23} & \underline{53.06} & \textbf{32.36} & \textbf{13.64} & \textbf{39.61} \\
        \midrule
        \rowcolor{gray!30} \multicolumn{12}{c}{\textbf{Close-source Models}} \\
        \textbf{GPT4o} & Text & 60.31 & 68.97 & 38.65 & \underline{50.12} & 63.04 & 71.14 & 73.37 & \underline{36.36} & \textbf{20.13} & \textbf{40.26} \\
        \textbf{GPT4o} & Image & 43.39 & 51.87 & 27.63 & 36.68 & 42.68 & 52.89 & 65.24 & 33.10 & 10.39 & 25.32 \\
        \textbf{GPT4o} & Image+Text & \underline{62.45} & \underline{69.01} & \underline{41.37} & 49.59 & \underline{65.91} & \underline{74.49} & 72.05 & 35.25 & \underline{14.29} & 30.52 \\
        \textbf{Gemini1.5-pro} & Text & 59.08 & 66.14 & 35.54 & 43.74 & 63.64 & 69.71 & 70.72 & 36.17 & 9.74 & 25.32 \\
        \textbf{Gemini1.5-pro} & Image & 50.37 & 57.62 & 25.42 & 33.76 & 41.52 & 50.23 & 67.22 & 36.26 & 7.79 & \underline{38.96} \\
        \textbf{Gemini1.5-pro} & Image+Text & 62.04 & 68.08 & 37.61 & 45.62 & 56.74 & 65.80 & \underline{74.80} & 36.26 & 9.09 & 24.03 \\ 
        \textbf{DeepSeek-R1} & Text & \textbf{70.91} & \textbf{79.45} & \textbf{70.31} & \textbf{72.54} & \textbf{82.71} & \textbf{84.62} & \textbf{79.55} & \textbf{42.59} & 7.14 & 32.14 \\
        \midrule
        \rowcolor{gray!30} \multicolumn{12}{c}{\textbf{Models with TreeThinker}} \\
        \textbf{GPT4o(TreeThinker)} & Text & \underline{64.50} & \underline{72.41} & \underline{53.34} & \textbf{65.08} & \underline{64.40} & \underline{75.67} & \underline{77.26} & \textbf{37.63} & \textbf{39.47} & \textbf{67.76} \\
        \textbf{GPT4o(TreeThinker)} & Image & 44.13 & 52.41 & 40.57 & 49.35 & 49.21 & 58.32 & 70.83 & 34.44 & 19.61 & \underline{67.32} \\
        \textbf{GPT4o(TreeThinker)} & Image+Text & \textbf{65.82} & \textbf{73.32} & \textbf{55.60} & \underline{64.28} & \textbf{66.31} & \textbf{77.42} & \textbf{79.45} & \underline{37.08} & \underline{33.55} & 65.13 \\ 
        \bottomrule
    \end{tabular}
    }
    \label{tab:model_performance}
\end{table*}

\subsection{Tree-based Reasoning}

Previous studies \cite{Irrelevant} have shown that LLMs are often distracted by irrelevant information. To address this issue, we prompt the model to decompose question into keywords and align them with the table headers, thereby helping it accurately identify the sub-table relevant to the question.

\paragraph{Decomposition.} Decomposing questions into more fine-grained components can simplify the reasoning process required and enhance the model’s performance. Therefore, we first prompt the model to decompose the question $Q$ into several keywords $K=[k_1,k_2,...,k_m]$, helping it focus on the most critical aspects of the question.

\paragraph{Aligning and Reasoning.} Once the question decomposition is finished, we instruct the model to align keywords with header tuples, enabling it to accurately pinpoint the headers relevant to the question. Specifically, given the keywords $K$ and the header tuples $H$, the goal of Aligning is to build a Keyword-Header Tree $H^{\prime}=select(T,Align_{LM}(T,k) > \theta), T\in H, k\in K$. The function $Align_{LM}(T,k)$ calculate the degree of matching between header tuple $T$ and keyword $k$. The $select()$ function then chooses the header $T$ whose matching degree with keyword $k$ exceeds the threshold $\theta$ and add them to the Keyword-Header Tree $H^{\prime}$. 
After that, we incorporate $H^{\prime}$ in prompts, guiding the LLM to retrieve essential information, e.g. relevant sub-tables, from tables with improved reasoning abilities.

Lastly, we also supplement a React-Style multi-round refinement strategy --- through multiple rounds of \textit{Thought}, \textit{Action} and \textit{Result}, ultimately outputting the final answer. Full prompts are shown in Table \ref{tab:first_round_tree} and \ref{tab:second_round_tree}. 

\section{Experiment}

\subsection{Experimental Setup}
\paragraph{Baselines.} We evaluated 25 models, including different modalities, with parameters ranging from 7B to 90B across four categories: (1) Close-source models, including GPT-o1 \citep{GPT-o1}, GPT-4o \citep{gpt}, Deepseek-R1-API\citep{deepseek}, Gemini-1.5-pro \citep{gemini}, QwQ \citep{QWQ} and Doubao-1.5-pro \citep{doubao}   . (2) Open-source language models, including Llama3s \citep{llama3}, Qwen2.5s \cite{qwen2.5}, Mistral \citep{mistral} and Deepseek-R1-Distalls \citep{deepseek}. (3) Open-source multimodal models, including LLaVA-1.5 \citep{llava}, mPLUG-owl3 \citep{mPLUG-Owl3}, mPLUG-owl2 \citep{mPLUG-Owl2}, and Llama3-Visions. (4) Table-oriented models, including TableGPT \citep{tablegpt}, TableLLMs \citep{tablebench}, TableLlama \citep{tablellama}, and Table-LLaVA \citep{multimodal_table_understanding}. 

\paragraph{Metrics.}
 For Fact Checking, Numerical Reasoning and Structure Comprehending, we refer to previous work \citep{MULTIHIERTT} using F1 and EM as score metric for these tasks. For chart generation, we calculate the ECR to assess code executability, extract y-axis values to compare with reference data, and compute PASS@1 to evaluate the pass rate. For Data Analysis, we calculate ROUGE-L \citep{rouge} to evaluate the objective similarity between generated and reference answers. Inspired by G-EVAL\citep{G-EVAL}, we also design an evaluation template using GPT-4o to evaluate the scores of model answers. A detailed evaluation prompt template can be found in Appendix~\ref{appendix:GPT-Eval Prompts}.

\subsection{Implementation Details}
\begin{figure}
    \centering
    \includegraphics[width=1\linewidth]{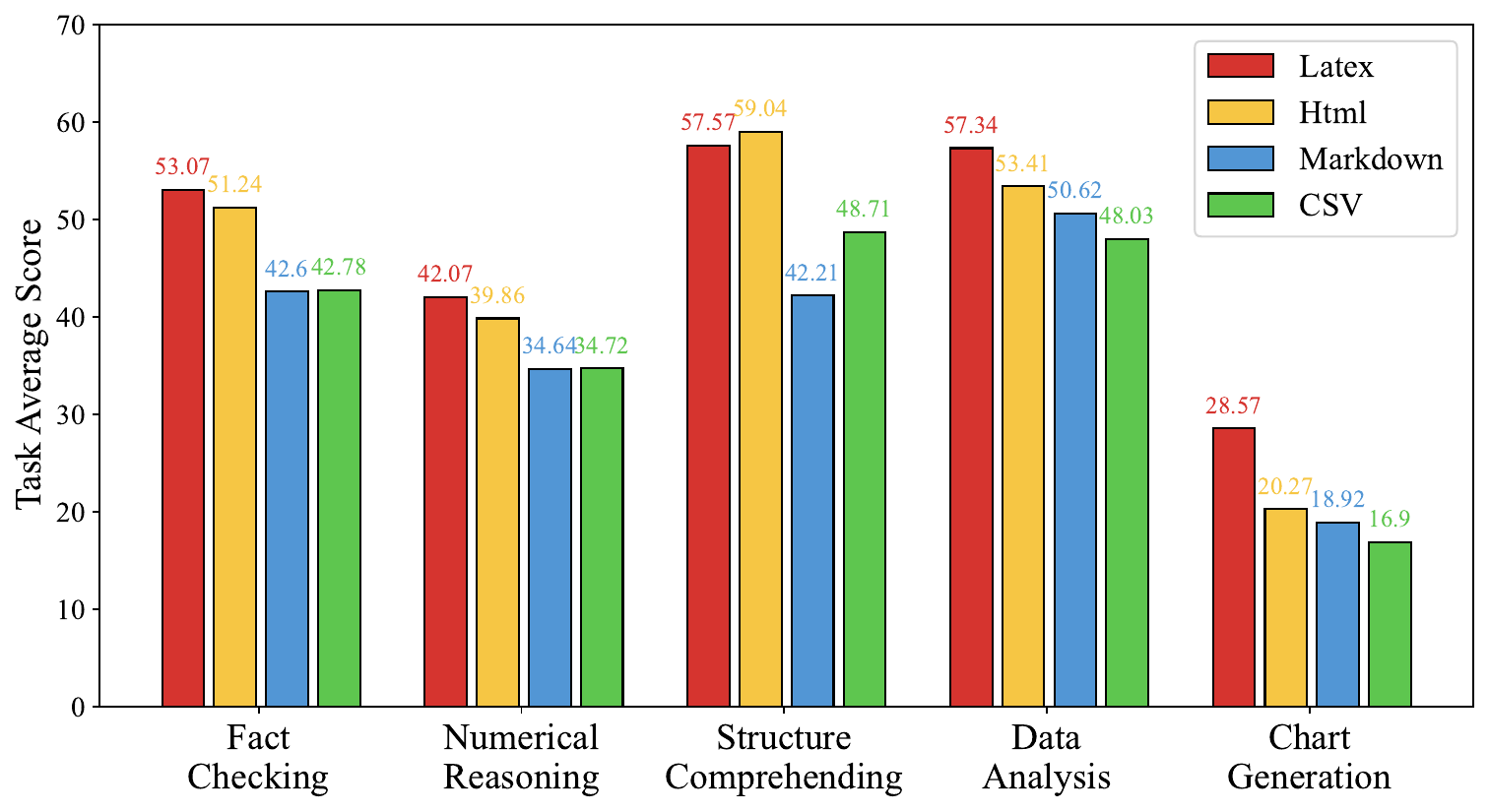}
    \caption{The comparison of average scores for different text formats across various tasks.} 
    \label{fig:formats_comparison}
    \vskip -0.15in
\end{figure}

We deploy open-source models on 8 H800 GPUs using the \textit{Transformer} library, while closed-source models are accessed through official APIs in accordance with their documentation. For table input formats, we tested GPT-4’s performance on LaTeX, HTML, Markdown, and CSV formats as text modality input, with LaTeX outperforming the others, as shown in the figure \ref{fig:formats_comparison}. Therefore we select Latex for text modality input. Meanwhile, we selected PNG format tables for visual modality input. Additionally, we set the model’s temperature to 0 to ensure deterministic outputs and configured the maximum output length to 4,096 tokens, balancing detail and efficiency. In order to focus on the hierarchical characteristics, we separate long tables from normal-size tables in experiment, but we still keep long tables in our dataset. For more details, please refer to the appendix~\ref{implement_detail}.

\subsection{Main Results}
\label{Main_Results}

\paragraph{\ding{172} Overall Performance.}
The results in Table ~\ref{tab:model_performance} highlight\textit{ the limitations of current LLMs in handling realistic hierarchical table analysis.} For our tasks, the EM metric of almost all models remains is relatively low, with the highest not exceeding 70. Plus, all models demonstrate remarkably low outcomes in Chart Generation, with code execution accuracy below 30. Interestingly, many Table-oriented models exhibit pronounced overfitting. However, TableGPT2 delivers the most impressive results among 7B-level models. For models with over 10B parameters, GPT-4o and LLama show comparable performance. It's worth mentioning that DeepSeek-R1 achieves the most outstanding results among all models. Although this might be attributed to its 671B MoE architecture, it also, to some extent, indicates the potential of reasoning for large models in addressing hierarchical structures.

\paragraph{\ding{173} The text modality outperforms the vision modality as an input.} On average, GPT-4o with text input outperforms image input by 15, while Gemini-Pro with text input exceeds image input by 10. Similarly, open-source MLLMs generally lag behind their backbone LLMs. Some table-oriented MLLMs like Table-LLaVa also perform worse than other textual ones. The performance of image inputs is not as good as that of text inputs, but they may serve as a complementary to text inputs.

\paragraph{\ding{174} Automatic tree reasoning significantly enhances structure understanding ability.} Compared to the baseline, GPT-4o with our TreeThinker method demonstrates consistent improvements across different input modalities, achieving the best performance in RealHiTBench. For example, GPT-4o in the Chart Generation, the PASS@1 for text+image input increased from 14.29 to \textbf{33.55}, marking a \textbf{134.7\% }improvement. Even for text-only input, TreeThinker enhances performance in NR, raising F1 from 36.68 to \textbf{49.35}. 

\paragraph{\ding{175} Too long table size still matters.}

\begin{table}[!ht]
\centering
\small
\caption{{The evaluation results of advanced models with Text and Image Inputs on RealHiTBench.}}
\renewcommand{\arraystretch}{1.1}
\resizebox{0.95\columnwidth}{!}{
\begin{tabular}{ccccc}
\toprule[0.5mm]
\multirow{2}{*}{\textbf{Model}} & \multirow{2}{*}{\textbf{Input}} & \multicolumn{3}{c}{\textbf{Table Tokens}} \\
\cmidrule[0.3mm](lr){3-5}
 &  & \textless{}\textbf{10K} & \textbf{10K-20K} & \textgreater{}\textbf{20K} \\
\midrule[0.3mm]
\textbf{GPT-4o} & Text & 56.45 & 40 & 30.77 \\
\textbf{GPT-4o} & Image & 42.54 & 21.83 & 13.48 \\
\textbf{Llama3} & Text & 50.9 & 39.14 & 27.78 \\
\textbf{Llama3V} & Image & 30.03 & 18 & 9.01 \\
\bottomrule[0.5mm]
\end{tabular}
}
\label{tab:table_size_evaluation}
\end{table}

While handling tables, we identify 127 tables with complex structures that are significantly larger, making it impossible for LLMs to take in the complete table within a single conversation during evaluation. We extracted the impact of different table sizes on the performance of various modal models, which indicates that long table sizes do affect models' table comprehension abilities, with a more pronounced impact on visual modality models. Given that long tables can exhibit unique complexities, we consider breaking down the large-scale content and inputting them into LLMs with multi-turn dialogues, which are likely to disrupt the tabular information of the tables, which is unacceptable. Therefore, we hope that future works can develop reasonable methods to fully utilize the value of these long tables, or that future LLMs can support the input requirements for enormous tables due to real usage requirements.

\paragraph{\ding{176} Each TreeThinker component enhances the model’s effectiveness.}

\begin{table}[!ht]
\centering
\caption{{The ablation results of TreeThinker when using GPT-4o and Llama3.3-70b-Instruct as base models.}}
\renewcommand{\arraystretch}{1.2}
\resizebox{1\columnwidth}{!}{
\begin{tabular}{lcccc}
\toprule[0.5mm]
\multirow{2}{*}{\textbf{Components of TreeThinker}} & \multicolumn{2}{c}{\textbf{GPT-4o}} & \multicolumn{2}{c}{\textbf{Llama3-70b}} \\
\cmidrule[0.3mm](lr){2-3} \cmidrule[0.3mm](lr){4-5}
 & Avg Score & Delta & Avg Score & Delta \\
\midrule[0.3mm]
\textbf{All} & 63.29 & 0 & 62.23 & 0 \\
\midrule[0.3mm]
\textbf{w/o Tree Generation} & 55.27 & \cellcolor{red!50} \textbf{-8.02} & 54.04 & \cellcolor{red!50} \textbf{-8.19} \\
\textbf{w/o Tree-based Reasoning} & 60.75 & \cellcolor{red!20} \textbf{-2.54} & 54.58 & \cellcolor{red!40} \textbf{-7.65} \\
\bottomrule[0.5mm]
\end{tabular}
}
\label{ablation}
\end{table}

As indicated in Table~\ref{ablation}, we conduct ablation studies on GPT-4o and Llama3-70b to assess the impact of various components on the performance of TreeThinker. We find that each component contributes to the model’s effectiveness, with Tree Generation playing a particularly crucial role in enhancing the model’s understanding of realistic complex hierarchical table.

\paragraph{Note:}We place more experimental results in the Appendix~\ref{more_experiments}.

\section{Conclusions and Future Directions}
In this paper, we establish a new benchmark for evaluating modern LLMs' ability to handle complex hierarchical tables.
Our results with 25 LLMs suggest that it is still challenging for LLMs to understand realistic complex hierarchical table analysis. 
Based on our full empirical findings, we anticipate further research in table analysis would consider some promising directions such as \textbf{(i)}-improving the visual structure understanding ability of MLLMs; \textbf{(ii)}-developing better tree-based structural reasoning techniques; \textbf{(iii)}-scaling LLMs to handle very long tables. 

\clearpage

\section*{Limitations}

We discover some limitations during benchmark construction. \textbf{(i)} We recruit and train 6 college students, who spent 540 hours each on data collection, processing, and annotation. We focus on the complexity of the tables and the accuracy of the annotations, so there is still room for improvement in terms of data volume. \textbf{(ii)} In the TreeThinker pipeline, we implement a multi-turn prompting approach, which significantly enhances the model's ability to understand table structures. However, this slows down the process of applying the TreeThinker pipeline, presenting a trade-off.

\section*{Ethical Considerations}

RealHiTBench, an open English benchmark that supports research in the field of table analysis. Our table data sources are all from well-known public data platforms on the Internet, and the data are fully usable. To ensure that our dataset does not contain any sensitive information, we establish strict review rules for annotators and set up 3 rounds of human checks throughout the annotation process, corresponding to the table, question, and answer, respectively. Therefore, we propose a benchmark that is clean and free from any ethical issues. More details about our dataset and annotation are in Section~\ref{sec:RealHiTBench} and Appendix~\ref{appendix:DatasetDetails} and \ref{appendix:AnnotationDetails}.

\section*{Acknowledgments}

This work is supported by NSFC under Grants (No. U24A201401) and partially supported by Pioneer R\&D Program of Zhejiang (No. 2024C01035).

\bibliography{RealHiTBench}

\appendix

\section{More Dataset Details}
\label{appendix:DatasetDetails}

In this section, we explain some details of our dataset. First, we investigate existing open platforms containing tables before collecting the data. Then, to build a unified dataset, we process the collected tables. Finally, we tally up the domains involved in the table content and topics to cover as many real-world application scenarios as possible.

\subsection{Data Sources}
\label{appendix:sources}

\begin{figure}
    \centering
    \includegraphics[width=1\linewidth]{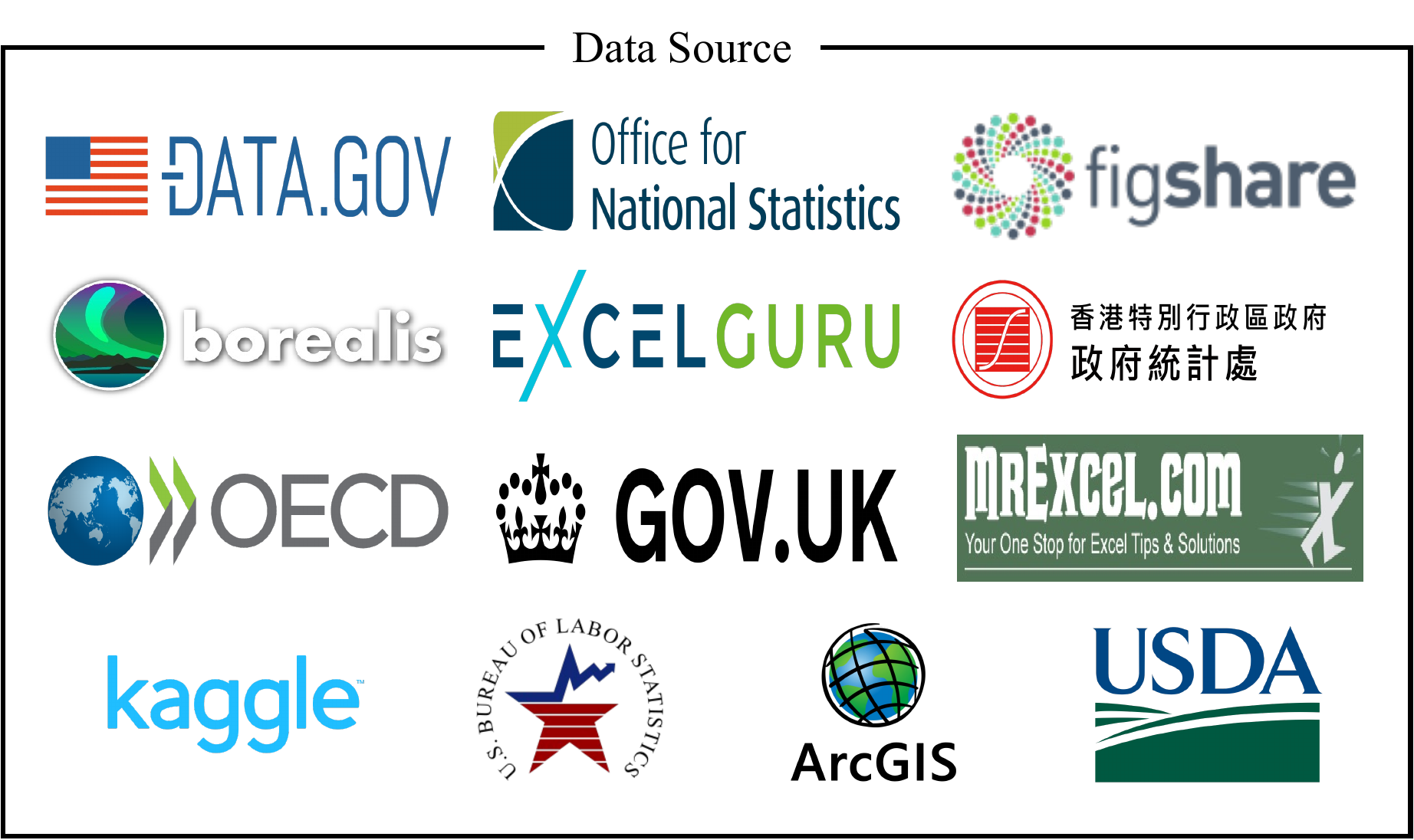}
    \caption{All data sources of tables in RealHiTBench.}
    \label{fig:source}
\end{figure}

The tables in our dataset are raw from 13 open platforms. We investigate these platforms in advance to ensure the authority and rationality of the data. Introduction of the sources is as following.

\paragraph{Kaggle\footnote{\url{https://www.kaggle.com/}}.} Kaggle is a valuable platform in the field of data science and machine learning. It hosts a variety of competitions that involve solving real-world problems. It also provides a vast repository of datasets across various domains. The data on Kaggle is generally considered to be of high quality and reliable, as it sourced from various places, including public datasets, partner datasets, and user-generated datasets.

\paragraph{USDA.gov\footnote{\url{https://www.usda.gov/}}.} The U.S. Department of Agriculture website provides a wealth of data related to the food, agriculture, and rural sectors. The data are of high quality and reliability, as they follow strict quality standards and are used to inform and enhance public and private decision-making on economic and policy issues. It is an authoritative source for agricultural data.

\paragraph{Catalog.data.gov\footnote{\url{https://catalog.data.gov/}}.} This is a catalog of datasets from the US government. It offers a wide range of data resources that are reliable and valuable for research and analysis in various fields. Data are collected and maintained by government agencies, ensuring their authority.

\paragraph{Borealisdata\footnote{\url{https://borealisdata.ca/}}.} Borealis, the Canadian Dataverse Repository, is a bilingual, multidisciplinary, secure, Canadian research data repository. It is supported by academic libraries and research institutions across Canada, providing reliable and valuable data for researchers. Data are managed according to the FAIR principles for scientific data management.

\paragraph{Arcgis\footnote{\url{https://hub.arcgis.com/}}.} Arcgis provides geospatial data and tools for mapping and analysis. Data are reliable and widely used in various industries, such as urban planning, environmental management, and transportation. It offers valuable insights through its geospatial capabilities.

\paragraph{ExcelGuru\footnote{\url{https://www.excelguru.ca/}}.} This platform focuses on Excel-related data and resources. Although not as comprehensive as some other platforms, it can provide valuable data and tips for Excel users, especially in the context of data analysis and management.

\paragraph{Bureau of Labor Statistics\footnote{\url{https://www.bls.gov/}}.} It is a principal statistical agency of the US Federal Statistical System. The data provided by the Bureau of Labor Statistics are highly reliable and authoritative, covering various aspects of labor economics and statistics, such as employment, unemployment, inflation, and productivity.

\paragraph{OECD\footnote{\url{https://www.oecd.org/}}.} The Organization for Economic Co-operation and Development (OECD) provides a wide range of data on economic, social, and environmental issues. Data are collected from member countries and are of high quality and reliability, making it a valuable resource for policy-making and research.

\paragraph{Gov.UK\footnote{\url{https://www.gov.uk/}}.} The UK government's website provides data on various topics, such as government reference data, statistics, and reports. The data is reliable and authoritative, as it is collected and maintained by government departments.

\paragraph{Censtatd.gov.hk\footnote{\url{https://www.censtatd.gov.hk/}}.} The Census and Statistics Department of Hong Kong provides data on Hong Kong's economy, population, and social affairs. The data is reliable and valuable for understanding the local situation and making informed decisions.

\paragraph{MrExcel\footnote{\url{https://www.mrexcel.com/}}.} Similar to ExcelGuru, MrExcel focuses on Excel - related content. It can offer valuable data and insights for Excel users, especially in terms of advanced Excel techniques and applications.

\paragraph{Office for National Statistics (UK)\footnote{\url{https://www.ons.gov.uk/}}.} It is the national statistics office of the UK. The data provided by the Office for National Statistics is highly reliable and authoritative, covering a wide range of topics, such as population, economy, and society.

\paragraph{Figshare\footnote{\url{https://figshare.com/}}.} Figshare is a repository for research data and outputs. It allows researchers to store, share, and publish their data, making it a valuable resource for the research community. The data on figshare is diverse and can be used for various research purposes.

\subsection{Data Formats}
\label{appendix:formats}
With the current development of large language models (LLMs) and Multimodal Large Language Models (MLLMs), it is necessary to explore the performance of textual and visual models and input formats.

\paragraph{Textual Formats.} In order to explore the performance of textual input formats, we transform our xlsx tables into various textual formats, including LaTeX, HTML, CSV, Markdown. We conduct a preliminary rationality analysis of these formats. Formats like CSV and Markdown use simple delimiters to represent structural information, which are unlikely to obtain complex tabular and supplementary information. And LaTeX and HTML seem capable to comprehensively exhibit abundant information in complex tables. Thus, we conduct experiment (shown in Figure~\ref{fig:formats_comparison}) to compare the performance of these textual formats, including CSV and Markdown to be rigorous. The result shows the LaTeX outperforms among these formats. Therefore, we choose LaTeX as our textual input format.

\paragraph{Visual Format.} Image-based representations have effectiveness on table reasoning, so we also explore the performance of visual format on the same benchmark as textual formats. We convert Excel tables into PNG image format and provide them to MLLMs along with corresponding prompts. However, we discover a shortcoming of image format of table representations. For some enormous tables in our collections, image format either cannot cover complete data in the table or makes the picture fuzzy in the vision of MLLMs, which makes it much difficult to understand long tables with image formats. Nevertheless, we conduct experiments both on text and image, finding that although the image modality performs worse than the text modality, the image can bring auxiliary improvements to the text in understanding complex tables.

\subsection{Data Domains} 
\label{appendix:domains}

\begin{figure}
    \centering
    \includegraphics[width=1\linewidth]{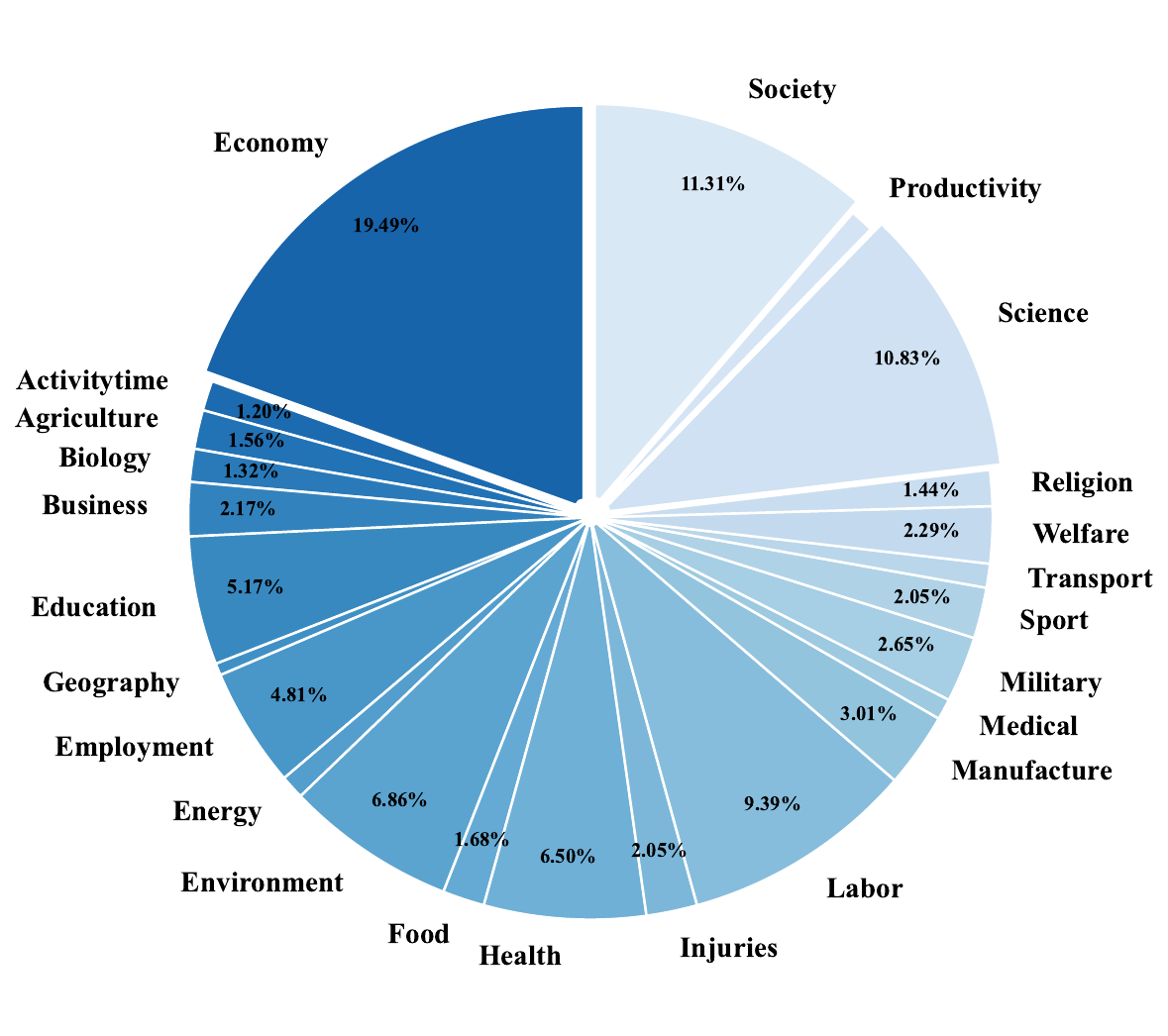}
    \caption{All domains of tables in RealHiTBench.}
    \label{fig:domain}
\end{figure}

We are devoted to constructing a realistic benchmark to reflect the complexity of tables in real scenarios. However, open-domain tables typically exhibit much simpler structures than the complex structures found in specific domains. Tables in specific domains commonly obtain characteristics that they contain domain-specific terms and typically not accompanied by domain-specific labeled data. Thus, it is not promising to conduct supervised fine-tuning on models for TableQA tasks~\cite{AIT-QA}. Therefore, we cover 24 domains (shown in Figure~\ref{fig:domain}) to align with real-world application scenarios. And we maintain a balance in the number of tables under each domain as much as possible.

\section{More Annotation Details}
\label{appendix:AnnotationDetails}

\subsection{Task types}
\label{appendix:task_types}

We are devoted to evaluate the performance of LLMs in table understanding and reasoning comprehensively. And table question answering (TableQA) is one of the most common applications about tables. Thus, referring to the previous work~\citep{tablebench}, we introduce 5 primary task types and a couple of more specific subtypes. Here we explain each type to avoid ambiguity.

\paragraph{Fact Checking.} Fact Checking in TableQA tasks is the process of verifying the accuracy and consistency of information within a table. It involves assessing whether the data presented in the table aligns with real-world facts or external sources of information. This category focuses on ensuring that the table's content is reliable and free from errors or inconsistencies.

\begin{itemize}
    \item \textbf{Multi-hop Fact Checking} This involves verifying the accuracy of a statement or answer that requires information from multiple cells or rows in a table. It is more complex than single-hop fact checking because it involves cross-referencing different parts of the table to ensure consistency and correctness. For example, verifying the relationship between two different data points in a table would require multi-hop fact checking.
    \item \textbf{Value-Matching} This involves checking whether a specific value or set of values in a table matches the requirements or conditions specified in a question or statement. It is straightforward and focuses on direct comparisons between the values in the table and the values mentioned in the question. For example, verifying the number of employees in a specific department would involve value-matching.
    \item \textbf{Inference-based Fact Checking} This involves verifying the accuracy of an answer that requires logical reasoning or inference based on the information provided in the table. It is more complex than value-matching because it requires drawing conclusions from the data in the table rather than simply matching values. For example, verifying the implications of a certain trend in the data would involve inference-based fact checking.
\end{itemize}

\paragraph{Numerical Reasoning.} Numerical Reasoning in TableQA tasks involves solving problems that require numerical calculations or analyses based on the data in a table. It encompasses tasks that involve arithmetic operations, comparisons, and other numerical computations to derive meaningful insights or answers from the tabular data.

\begin{itemize}
    \item \textbf{Multi-hop Numerical Reasoning} This involves solving numerical problems that require information from multiple cells or rows in a table. It is more complex than single-hop numerical reasoning because it involves cross-referencing different parts of the table to derive the final answer. For example, calculating the total sales of a product across multiple regions would require multi-hop numerical reasoning.
    \item \textbf{Counting} This involves determining the number of occurrences of a specific value or condition in a table. It is straightforward and focuses on counting the instances that meet the specified criteria. For example, counting the number of employees with a specific job title would involve counting.
    \item \textbf{Sorting} This involves arranging the data in a table in a specific order based on a particular attribute or condition. It is used to organize the data in a way that makes it easier to analyze or answer questions. For example, arranging sales data in descending order to identify the top three highest sales figures would involve sorting.
    \item \textbf{Comparison} This involves comparing the values of different cells or rows in a table to determine their relative magnitudes or relationships. It is used to answer questions that require identifying the highest, lowest, or other comparative relationships between data points. For example, determining which product has the highest sales would involve comparison.
    \item \textbf{Calculation} This involves performing mathematical operations on the data in a table to derive a final answer. It is used to answer questions that require arithmetic operations such as addition, subtraction, multiplication, or division. For example, calculating the average sales per region would involve calculation.
\end{itemize}

\paragraph{Data Analysis.} Data Analysis in TableQA tasks is the process of extracting meaningful insights and patterns from the data in a table. It involves tasks such as identifying trends, relationships, and anomalies within the data. This category focuses on understanding the data's underlying structure and using it to answer more complex questions or make predictions.

\begin{itemize}
    \item \textbf{Rudimentary Analysis} This involves performing basic analytical tasks on the data in a table, such as identifying patterns, trends, or simple relationships between data points. It is straightforward and focuses on providing a basic understanding of the data. For example, identifying the most common job title in a department would involve rudimentary analysis.
    \item \textbf{Summary Analysis} This involves providing a concise summary of the data in a table, highlighting key findings or insights. It is used to answer questions that require a brief overview of the data rather than detailed calculations or in-depth analysis. For example, providing a summary of sales performance across different regions would involve summary analysis.
    \item \textbf{Predictive Analysis} This involves using the data in a table to make predictions about future trends or outcomes. It is more complex and often involves statistical or machine learning techniques to model the data and make predictions. For example, making a prediction of future sales based on historical data would involve predictive analysis.
    \item \textbf{Exploratory Analysis} This involves examining the data in a table to identify patterns, relationships, or anomalies that may not be immediately obvious. It is used to answer questions that require a deeper understanding of the data and often involves visualizations or statistical techniques to explore the data. For example, analyzing the relationship between employee performance and job satisfaction would involve exploratory analysis.
    \item \textbf{Anomaly Analysis} This involves identifying and analyzing data points in a table that deviate significantly from the norm or expected values. It is used to answer questions that require identifying outliers or unusual data points that may indicate errors or special cases. For example, identifying any unusual sales figures in a table would involve anomaly analysis.
\end{itemize}

\paragraph{Chart Generation.} Chart Generation in TableQA tasks involves creating visual representations of the data in a table to make it more intuitive and easier to understand. It includes generating various types of charts and graphs, such as line charts, bar charts, scatter charts, and pie charts. This category helps users quickly grasp the key information and trends in the data.

\begin{itemize}
    \item \textbf{LineChart Generation} This involves creating a line chart to visually represent the data in a table, typically to show trends or changes over time. It is used to answer questions that require a clear and intuitive display of how data points change over a period. For example, visualizing sales trends over the past year would involve line chart generation.
    \item \textbf{BarChart Generation} This involves creating a bar chart to visually represent the data in a table, typically to compare the values of different categories or groups. It is used to answer questions that require a clear and intuitive display of the relative magnitudes of different data points. For example, visualizing sales figures across different regions would involve bar chart generation.
    \item \textbf{ScatterChart Generation} This involves creating a scatter chart to visually represent the data in a table, typically to show the relationship between two variables. It is used to answer questions that require a clear and intuitive display of the correlation or pattern between two sets of data points. For example, visualizing the relationship between employee performance and job satisfaction would involve scatter chart generation.
    \item \textbf{PieChart Generation} This involves creating a pie chart to visually represent the data in a table, typically to show the proportion of each category or group relative to the whole. It is used to answer questions that require a clear and intuitive display of the distribution of data points across different categories. For example, visualizing the distribution of employees across different departments would involve pie chart generation.
\end{itemize}

\paragraph{Structure Comprehending.} Structure comprehending involves providing a new table that is created by exchanging some complex parts of the source table and then asking the model the same question with two similar tabular inputs. The goal is to evaluate whether the model can accurately comprehend the structural changes and provide the correct answer. This type of task is particularly challenging because it requires the model to not only understand the data but also the intricate relationships and hierarchies within the table structure.

\subsection{QA Annotation}

    \begin{figure*}
        \centering
        \includegraphics[width=1\linewidth]{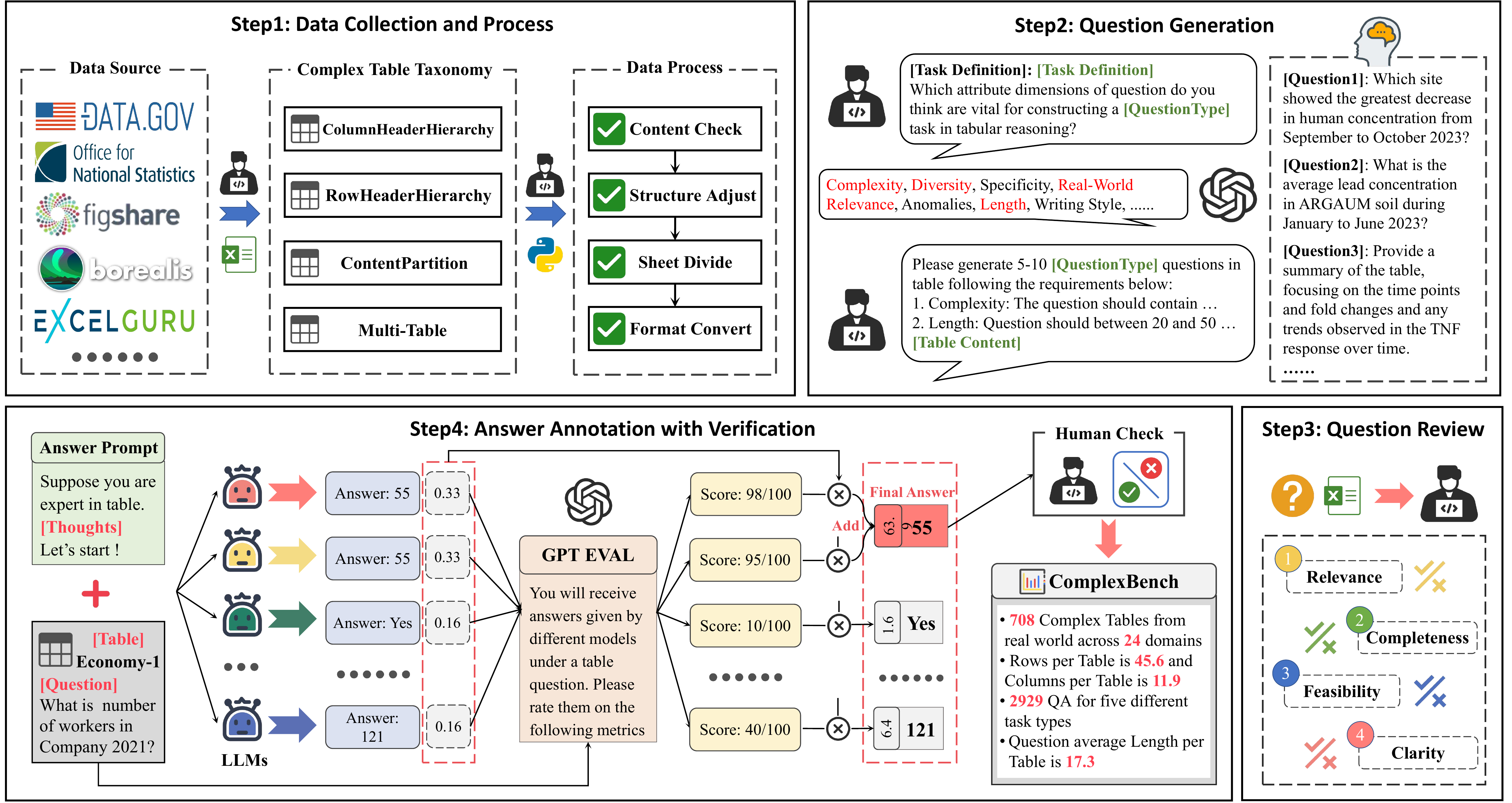}
        \caption{Process diagram of our dataset construction process.} \label{dataset_construction}
    \end{figure*}

Complex tables alone cannot fully present the challenges in RealHiTBench. Therefore, we meticulously design an annotation process that is suited for question answering tasks on complex tables. Specifically, we apply LLMs to initial question answering (QA) generation to increase annotation efficiency, and we keep the accuracy of annotations by careful multi-turn human check. The complete process diagram is depicted as Figure~\ref{dataset_construction}, and introduced as below: 

\paragraph{Data Collection and Process.} Initially, we collect tabular data from various sources. Then we divide the tables into different categories according to their structures. To tables with complex structures, we process tables before accepting them. Specifically, first we check whether table content contains no improper information. When content is done, we adjust the unrestricted structures to emphasize the complex parts. Because we collect tables as xlsx format, which allows a file to contain multiple sheets, we conduct sheet division to keep the needed table only. Finally, we convert the source table into chosen textual format LaTeX and visual format PNG (The reasons to choose these two formats have been clarified in Appendix~\ref{fig:formats_comparison}). Then a qualified table is added into our dataset.

\paragraph{Question Generation.} In order to display the complexity in our dataset comprehensively, we introduce LLM to assist in generating annotations. Before generating questions, we interact with GPT to acquire some inspirations about the attribute dimensions of question that are vital for constructing tasks in tabular reasoning~\citep{attribute}. We select some points from the GPT answers, including \textit{Complexity}, \textit{Diversity}, \textit{Real-World Relevance}, and \textit{Length}. Then, we design our prompt templates from these dimensions (shown in Table ~\ref{tab:prompt_q_fact_checking}). Moreover, we also design different prompt templates for each task type to ensure highlighting the characteristics of each type. Eventually, we combine the prompt template with the exact question type and table content (xlsx format to simulate human intuitive vision) as input for GPT, acquiring the generated answer.

\paragraph{Question Review.} Referring to the previous work, the LLM-assisted annotation is likely to lead to the bias or instability of benchmark~\citep{LLM_is_not_annotator}. Therefore, we set question review step to ensure the rationality of questions generated by GPT. We set four attributes of a reasonable question for annotators. Then we provide the tables and questions along with these attributes as review standard to annotators to confirm rationality or adjust the questions.

\paragraph{Answer Annotation with Verification.} Similarly, we also design a prompt template for answer generation (shown in Table~\ref{tab:prompt_a_fact_checking}). We combine the prompt template, table content (LaTeX format to simulate careful observation), and question as initial input for models. Then we feed these input to various models and acquire some answers. At the same time, the occurrence times of each answer can be automatically counted as their frequency. Then we input all the generated answers along with initial answers into GPT-Eval module, and gain the score of each answer. We multiply each score by the occurrence frequency of the answer to obtain a final score. The answer with the highest final score becomes the candidate answer. Finally, the candidate answer is provided to annotators for human check and the corrected final answer can be filled into the annotation field.

\subsection{QA prompt}

\paragraph{Question Prompt.} We introduce question prompts to generate diverse questions with GPT. Considering the uniqueness of different task types, we meticulously design different prompts for each type according to corresponding attribute dimensions (have been explained above). we provide our complete prompts from Table~\ref{tab:prompt_q_fact_checking} to Table~\ref{tab:prompt_q_chart_generation}.

\paragraph{Answer Prompt.} We also introduce answer prompts to generate various questions according to different task type with GPT. Apart from the fundamental prompt parts, such as \textit{Role Play} and \textit{Output Control}, we apply \textbf{COT}~\citep{COT} to the answer prompt to improve the capability of understanding. All answer prompts are presented from Table~\ref{tab:prompt_a_fact_checking} to Table~\ref{tab:prompt_a_chart_generation}. Notably, because the answer content and formats obviously vary among different subtypes of Data Analysis, we design detailed answer prompt for each subtype (Here we just display the prompt of rudimentary analysis.).

\paragraph{Tree-Based Answer Prompt.} We introduce TreeThinker~\ref{TreeThinker}, a two-round prompt pipeline that enhances the model's ability in table understanding, whose performance is shown the experimental results above. Here, we provide the detailed prompts in Table~\ref{tab:first_round_tree} and Table~\ref{tab:second_round_tree}, respectively.

\section{More Implementation Details}
\label{implement_detail}
\begin{table*}[!ht]
\centering
\caption{{The more results of advanced models with Text and Image Inputs on RealHiTBench.}}
\renewcommand{\arraystretch}{1.3}
\resizebox{1\textwidth}{!}{
\begin{tabular}{cccccccccccc}
\toprule[0.5mm]
 &  & \multicolumn{2}{c}{\textbf{Fact Checking}} & \multicolumn{2}{c}{\textbf{Numerical Reasoning}} & \multicolumn{2}{c}{\textbf{Structure Comprehending}} & \multicolumn{2}{c}{\textbf{Data Analysis}} & \multicolumn{2}{c}{\textbf{Chart Generation}} \\
\cmidrule[0.3mm](lr){3-4} \cmidrule[0.3mm](lr){5-6} \cmidrule[0.3mm](lr){7-8} \cmidrule[0.3mm](lr){9-10} \cmidrule[0.3mm](lr){11-12}
\multirow{-2}{*}{\textbf{Model}} & \multirow{-2}{*}{\textbf{Method}} & F1 & EM & F1 & EM & F1 & EM & ROUGE & GPT-EVAL & ECR & PASS \\
\midrule[0.3mm]
QwQ-32B & Text & 43.13 & 28.95 & 33.57 & 17.94 & 37.96 & 12.99 & 20.86 & 68.66 & 31.25 & 12.50 \\
Doubao-1.5-pro-32k & Text & \textbf{66.34} & \textbf{59.65} & \textbf{53.77} & \textbf{45.40} & \textbf{72.45} & \textbf{65.90} & \textbf{36.01} & \textbf{75.21} & 25.97 & \textbf{14.94} \\
DeepSeek-R1-Distill-Qwen-1.5B & Text & 17.75 & 11.18 & 15.74 & 7.13 & 19.93 & 9.92 & 16.24 & 38.02 & 14.94 & 0 \\
DeepSeek-R1-Distill-Qwen-7B & Text & 34.53 & 28.10 & 24.96 & 17.64 & 28.10 & 19.08 & 25.19 & 54.39 & \textbf{39.61} & 6.49 \\
DeepSeek-R1-Distill-Llama-8B & Text & 33.29 & 22.43 & 24.64 & 15.18 & 26.66 & 14.76 & 26.98 & 54.55 & 12.99 & 5.19 \\
mPLUG-Owl2-7B & Image & 7.23 & 3.46 & 4.72 & 1.15 & 14.21 & 9.14 & 12.13 & 25.46 & 1.53 & 0 \\
Qwen2-VL-7B-Instruct & Image & 38.34 & 28.50 & 22.88 & 9.34 & 55.99 & 45.29 & 24.30 & 37.41 & 16.23 & 5.84 \\ 
Qwen2-VL-7B-Instruct & Image+Text & 45.19 & 32.79 & 24.97 & 11.54 & 60.42 & 48.60 & 25.39 & 39.69 & 25.32 & 9.09 \\
\bottomrule[0.5mm]
\end{tabular}
}
\label{more_result}
\end{table*}

\begin{table}[!ht]
\centering
\small
\caption{{The evaluation results of reasoning models on RealHiTBench.}}
\renewcommand{\arraystretch}{1.3}
\resizebox{1\columnwidth}{!}{
\begin{tabular}{ccccccc}
\toprule[0.5mm]
\multirow{2}{*}{\textbf{Models}} & \multirow{2}{*}{\textbf{Method}} & \multicolumn{5}{c}{\textbf{Avg Score}} \\
\cmidrule[0.3mm](lr){3-7}
 &  & FC & NR & SC & DA & CG \\
\midrule[0.3mm]
\multirow{2}{*}{\textbf{GPT-o1}} & CoT & 70.61 & 63.08 & 78.72 & 54.02 & 28.73 \\
 & TT & 75.83 & 65.63 & 81.92 & 56.50 & 66.67 \\
 \midrule[0.3mm]
\multirow{2}{*}{\textbf{DeepSeek-R1}} & CoT & 75.18 & 71.43 & 83.67 & 61.07 & 19.64 \\
 & TT & 79.18 & 71.23 & 84.61 & 68.94 & 61.91 \\
 \midrule[0.3mm]
\multirow{2}{*}{\textbf{QwQ-32B}} & CoT & 41.79 & 23.12 & 28.83 & 47.39 & 13.27 \\
 & TT & 47.77 & 41.235 & 47.39 & 49.69 & 35.72 \\
\bottomrule[0.5mm]
\end{tabular}
\label{reasoning_result}
}
\end{table}

\subsection{Omit Long Tables}
\label{appendix:omit_long_tables}

\begin{table*}[!ht]
\centering
\caption{{The results of TreeThinker when using Llama3.1-8B-Instruct and Llama3.3-70b-Instruct as the base models.}}
\renewcommand{\arraystretch}{1.3}
\resizebox{1\textwidth}{!}{
\begin{tabular}{cccccccccccc}
\toprule[0.5mm]
 &  & \multicolumn{2}{c}{\textbf{Fact Checking}} & \multicolumn{2}{c}{\textbf{Numerical Reasoning}} & \multicolumn{2}{c}{\textbf{Structure Comprehending}} & \multicolumn{2}{c}{\textbf{Data Analysis}} & \multicolumn{2}{c}{\textbf{Chart Generation}} \\
\cmidrule[0.3mm](lr){3-4} \cmidrule[0.3mm](lr){5-6} \cmidrule[0.3mm](lr){7-8} \cmidrule[0.3mm](lr){9-10} \cmidrule[0.3mm](lr){11-12}
\multirow{-2}{*}{\textbf{Model}} & \multirow{-2}{*}{\textbf{Method}} & F1 & EM & F1 & EM & F1 & EM & ROUGE & GPT-EVAL & ECR & PASS \\
\midrule[0.3mm]
 & CoT & 44.93 & 30.32 & 27.21 & 14.53 & 50.8 & 35.9 & 32.25 & 60.12 & 13.64 & {\color[HTML]{1F2329} 4.55} \\
\multirow{-2}{*}{\textbf{Llama3.1-8B-Instruct}} & TT & 47.4 & 39.85 & 37.69 & 29.18 & 52.17 & 44.27 & {\color[HTML]{1F2329} 34.62} & {\color[HTML]{1F2329} 61.69} & 35.06 & 27.27 \\
\midrule[0.3mm]
 & CoT & 64.53 & 53.08 & 48.99 & 36.58 & {\color[HTML]{1F2329} 68.93} & {\color[HTML]{1F2329} 55.81} & 27.98 & 52.26 & 50.65 & {\color[HTML]{1F2329} 24.03} \\
\multirow{-2}{*}{\textbf{Llama3.3-70B-Instruct}} & TT & 69.94 & 63.93 & 61.69 & 54.6 & 73.22 & 64.89 & 37.38 & 70.57 & 76.62 & \underline{35.71} \\
\bottomrule[0.5mm]
\end{tabular}
}
\label{treethinker_open}
\end{table*}

We collect as much difficult tables as we can so that some enormous tables are in our dataset. However, considering the structural complexity, we are also devoted to multimodal exploration. Unfortunately, it seems impossible to input some enormous tables in realistic applications into MLLMs because of the oversize. Therefore, we omit long tables in our experiment, but we still keep them for future use.

\subsection{Chart Generation}

As shown in Table~\ref{tab:model_performance}, Chart Generation is too difficult, leading to the PASS@1 results for some models being 0. We investigate executable python codes generated by models, finding that \textit{DataFrame}, a common python module, exists in most of these codes. However, we find it difficult for python module \textit{DataFrame} to represent hierarchical table structures in detail, obscuring models from meaningful hierarchical information. Finally, non-functional codes are generated.

\subsection{GPT-Eval Prompts}
\label{appendix:GPT-Eval Prompts}

As demonstrated above, we utilize GPT-Eval to assess the reliability of answers to some open questions (in the Data Analysis task). Thus, we also design a series of prompts to evaluate different task types. The example prompt is presented in Table~\ref{tab:GPT-Eval_prompt}, which is a representative of the GPT-Eval prompts due to the similarity of these subtypes.

\section{More Experiments}
\label{more_experiments}

\subsection{Experiments on Reasoning Models.}
We evaluated the performance of the most advanced inference models, GPT-o1, DeepSeek-R1, and QWQ on RealHitBench. Considering the high computational cost of GPT-o1 and the recent instability of DeepSeek’s server due to a surge in traffic, we randomly selected 500 samples from benchmark to assess the reasoning models’ ability to understand realistic hierarchical complex tables, with the results shown in the figure~\ref{reasoning_result}. Compared to general-purpose models, inference-based models perform better in handling complex tables. Moreover, our proposed TreeThinker further enhances their ability to process tabular data.

\subsection{Different size and complexity table.}
We evaluated the overall performance of the model under varying table token counts and different levels of table complexity, as shown in the figure~\ref{tab:complexity_comparison}. The results indicate that as the number of tokens increases and the table complexity rises, the model’s performance gradually declines. Moreover, when these two factors are combined, the challenge for the model becomes even greater.

\subsection{TreeThinker on open-source models.}
\label{appendix:TreeThinker}
As shown in the table \ref{treethinker_open}, we evaluated TreeThinker on open-source models, specifically Llama3.1-8B-Instruct and Llama3.3-70B-Instruct, across multiple reasoning and analysis tasks. The results demonstrate consistent improvements over the CoT baseline in all tasks. Notably, TreeThinker significantly enhances model's ability in Chart Generation, with ECR rising from 50.65 to 76.62 and PASS from 24.03 to 35.71 in Llama3.3-70B-Instruct. These results highlight TreeThinker’s effectiveness in enhancing open-source models across a diverse range of tasks. Regardless of model size, TreeThinker consistently improves performance in various reasoning and analytical tasks, demonstrating its effectiveness in boosting the comprehension and processing of complex hierarchical data.

\section{Case Study}

\subsection{Complex Tables Examples}

To better illustrate the complexity of each table structure, we provide examples from our dataset, including tables and corresponding question answering (QA) pairs. By the way, to make the table structure more intuitive, we appropriately process the table in the dataset to highlight its complex structures.

\begin{figure*}
    \centering
    \includegraphics[width=1\linewidth]{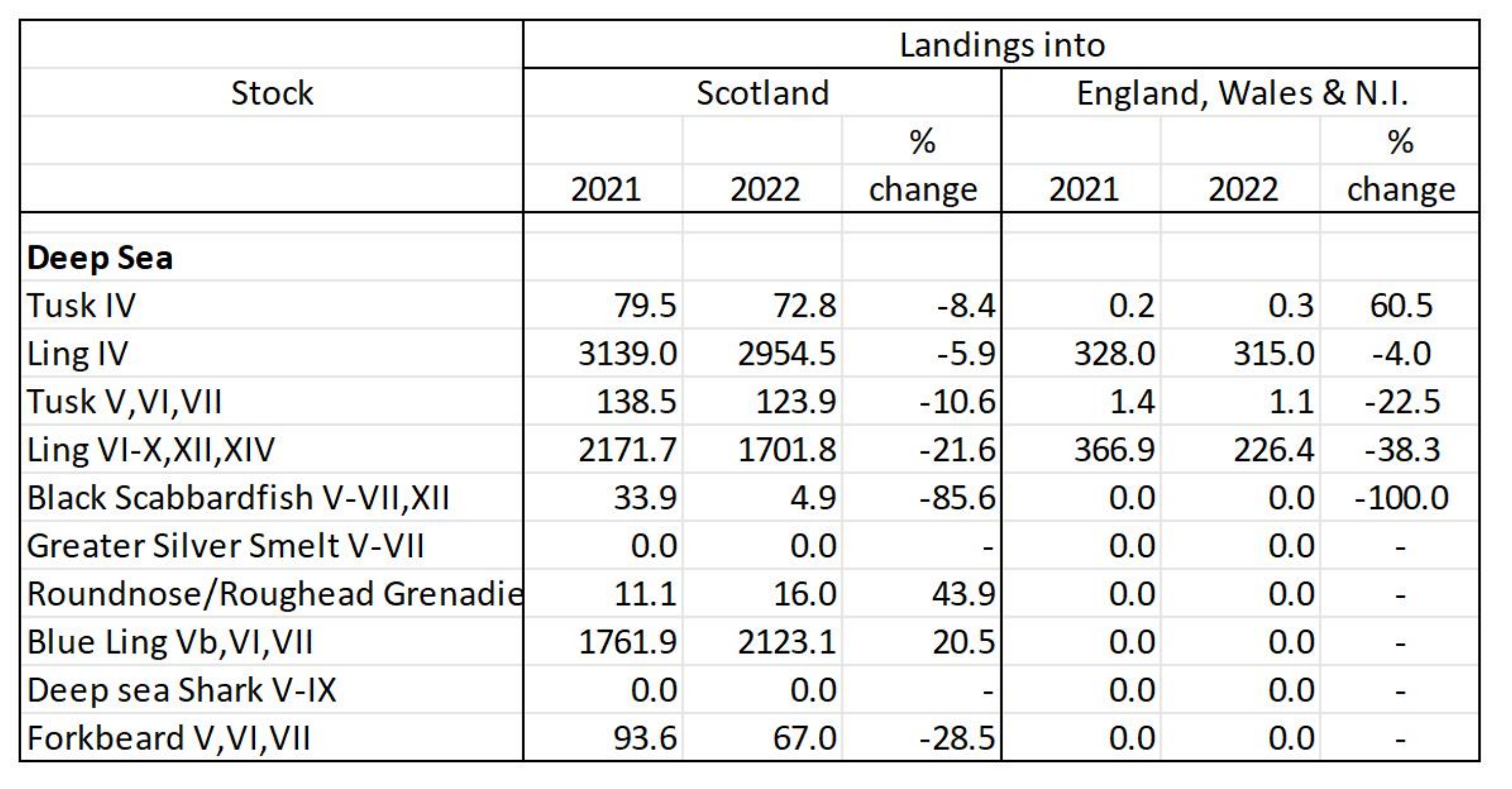}
    \caption{The \textbf{Hierarchical Column Header} sample in RealHiTBench.}
    \label{fig:Hierarchical_column_header_sample}
\end{figure*}

\paragraph{Hierarchical Column Header.} The specific example is presented in Figure~\ref{fig:Hierarchical_column_header_sample}. Merged cells, such as 'Landings into' and 'Scotland', present the Hierarchical Column Header structure.

\begin{figure*}
    \centering
    \includegraphics[width=1\linewidth]{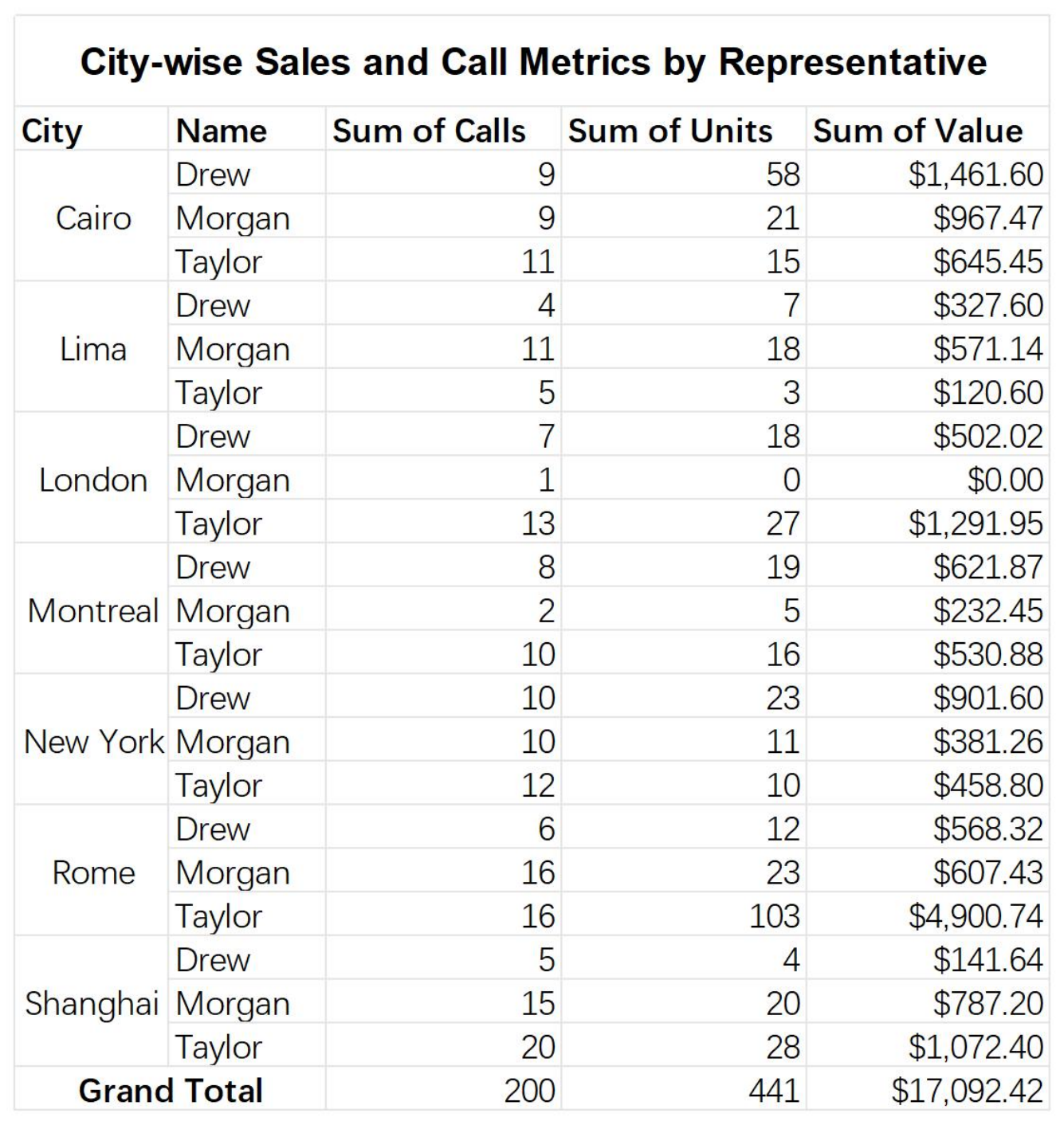}
    \caption{The \textbf{Hierarchical Row Header} sample in RealHiTBench.}
    \label{fig:Hierarchical_row_header_sample}
\end{figure*}

\paragraph{Hierarchical Row Header.} Hierarchy structure in row headers is depicted in Figure~\ref{fig:Hierarchical_row_header_sample}, which shows the type that the classification is not in the same column. The other type, classification shown in the same column, is displayed in Figure~\ref{table_sample}.

\begin{figure*}
    \centering
    \includegraphics[width=1\linewidth]{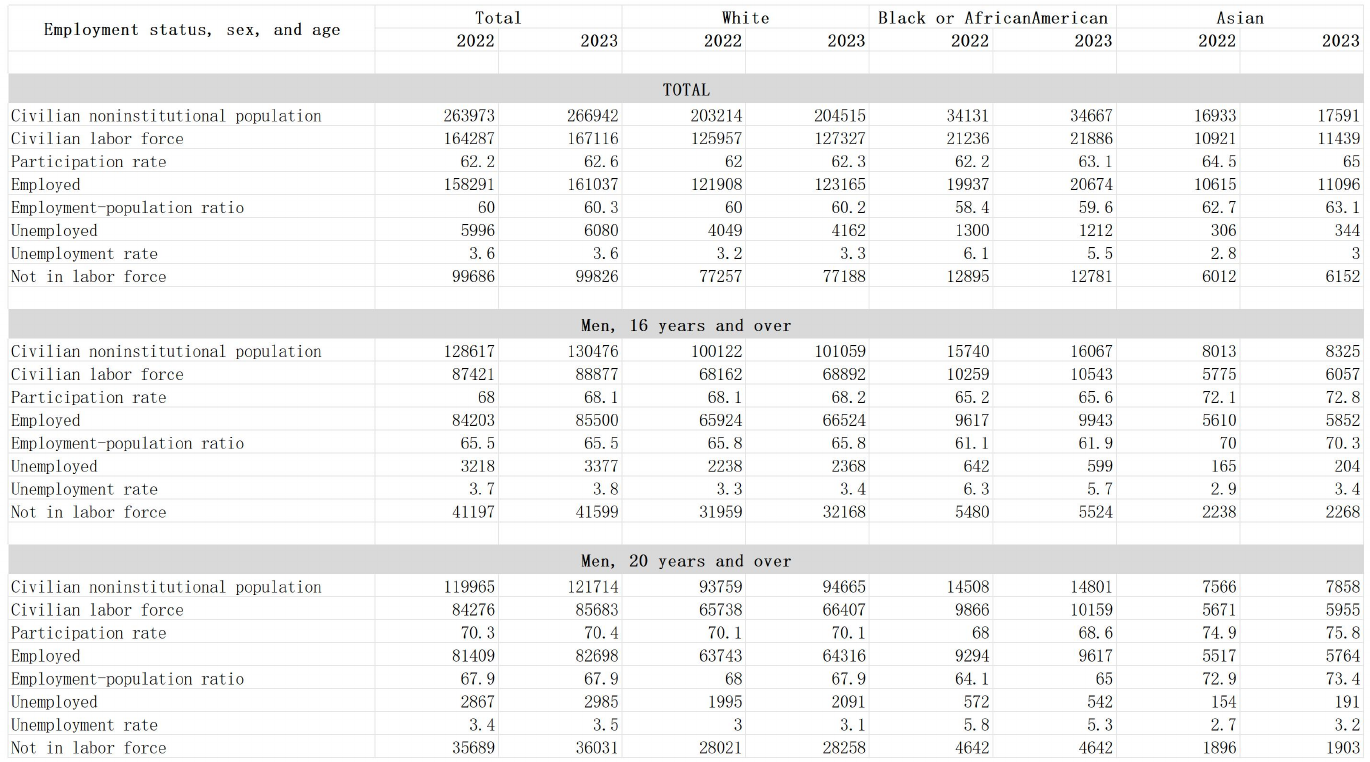}
    \caption{The \textbf{Nested Sub Tables} sample in RealHiTBench.}
    \label{fig:Nested_sub_tables_sample}
\end{figure*}

\paragraph{Nested Sub-Tables.} Just as Figure~\ref{fig:Nested_sub_tables_sample}, the whole table content is divided into several sub-tables by merged cells spanning the full width of the table. Notably, the content in sub-table 'TOTAL' is the superset of the other parts, while the data in area 'Men, 16 years and over' are the superset of the segments below. This trick sometimes causes ambiguities.

\begin{figure*}
    \centering
    \includegraphics[width=1\linewidth]{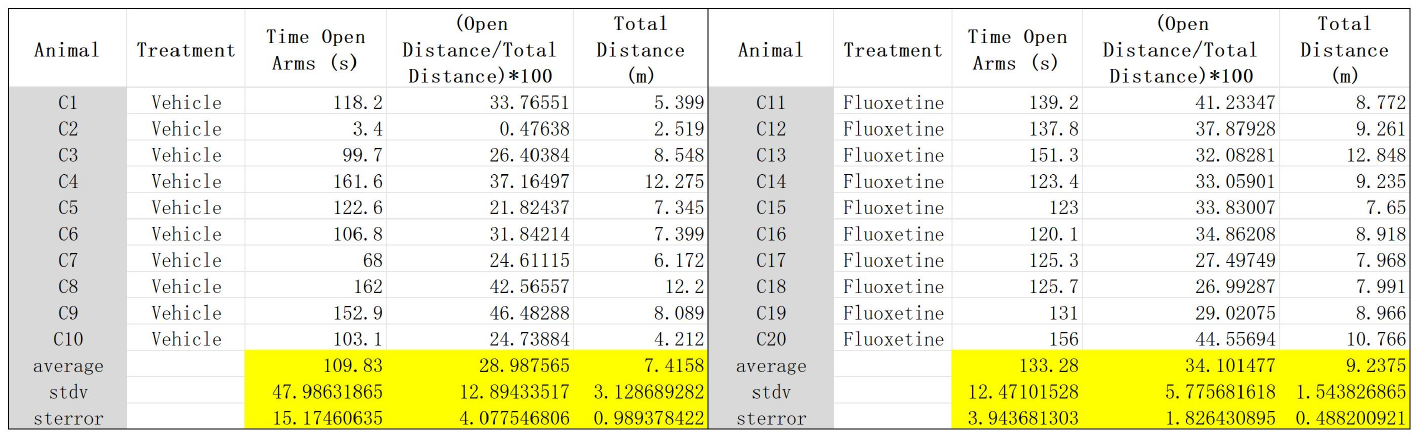}
    \caption{The \textbf{Multi-Table Join} sample in RealHiTBench.}
    \label{fig:Multi-Table_Join}
\end{figure*}

\paragraph{Multi-Table Join.} We prepare Figure~\ref{fig:Multi-Table_Join} as the example of Multi-Table Join. Actually, columns headers of the table consist of 5 cells, which look like 10 cells though. The table-maker present the tabular data like this is to create intuitive comparison between 'C1-C10' and 'C11-C20'.

\begin{table}[!ht]
  \caption{Comparison between RealHiTBench and other existing related benchmarks across two aspects: table size and complexity score.}
  \label{tab:complexity_comparison}
  \centering
  \renewcommand{\arraystretch}{1.2}
  \begin{threeparttable}
  \resizebox{\linewidth}{!}{
  \begin{tabular}{lcccc}
    \toprule
    Benchmark & TableBench & MULTIHIERTT & HiTab & RealHiTBench \\
    \midrule
    \rowcolor{gray!20} \multicolumn{5}{l}{\textbf{Table Size}} \\
    \hspace{2pt} 70th Pctl Row & 19 & 10 & 23 & \textbf{69} \\
    \hspace{2pt} 70th Pctl Column & 7 & 5 & 13 & \textbf{16} \\
    \hspace{2pt} 70th Pctl Cell Length & 7.52 & 11.92 & 7.30 & \textbf{9.59} \\
    \midrule
    \rowcolor{gray!20} \multicolumn{5}{l}{\textbf{Complexity Score}} \\
    \hspace{2pt} Header Hierarchy & 27.30 & 38.62 & 57.91 & \textbf{60.55} \\
    \hspace{2pt} Content Partition & 19.45 & 32.16 & 47.10 & \textbf{57.54} \\
    \hspace{2pt} Implicit Multi-Table & 12.34 & 21.52 & 33.98 & \textbf{43.48} \\
    \hspace{2pt} Supplement Information & 32.59 & 37.26 & 55.64 & \textbf{69.86} \\
    \bottomrule
  \end{tabular}
  }
  \begin{tablenotes}  
        \footnotesize  
        \item[1] 70th Pctl: 70th Percentile
      \end{tablenotes}        
    \end{threeparttable}
\end{table}

\begin{figure}
    \centering
    \includegraphics[width=1\linewidth]{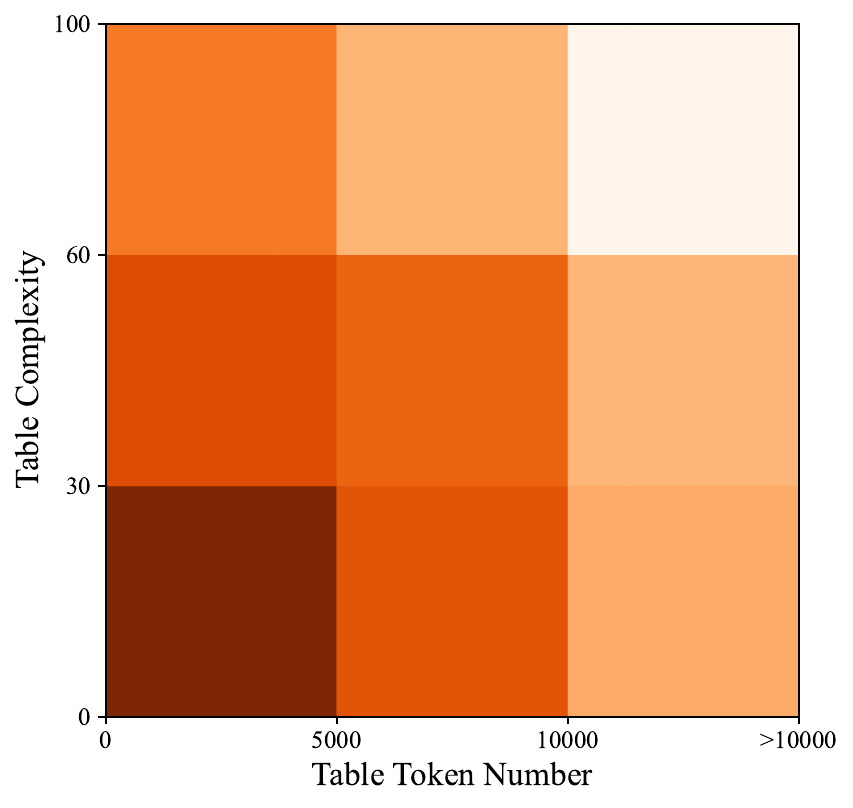}
    \caption{Different complexity and tokens of tables comparsion in RealHiTBench.}
    \label{fig:complex_token}
\end{figure}

\begin{table*}[!ht]
    \centering
    \caption{The complete question prompt of \textbf{Fact Checking}. Fill the corresponding content into the '...' positions after '[]' field.}
    \label{tab:prompt_q_fact_checking}
    \begin{tabular}{m{15cm}}
    
    \hline
     \textbf{Question Generation Prompt for Fact Checking} \\
    \midrule
     \# \textit{Role play} \\
     Suppose you are an expert in question annotation and your task is to generate high-quality and diverse questions based on a given task description and tabular data. \\

     \hline
     
     \# \textit{Task Descriptions} \\
     Fact checking task in tabular data involves locating relevant data from a table based on a question posed by a user and performing the necessary lookups, calculations, or comparisons to verify or generate an accurate answer. The task typically involves extracting keywords from the question, locating specific rows and columns of the table, and performing calculations, aggregations, or conditional filtering of the data, if necessary, to produce a result that meets the requirements of the question.\\

     \hline

     \# \textit{Generation Restrictions} \\
     To generate 5 questions based on the given task description and tabular data, give due consideration to the following aspects: \\
    1.Complexity: Include a range of problem complexities to suit different levels of fact-checking skill. Simple problems should involve straightforward calculations or direct comparisons, like checking totals or averages, while complex questions might involve multi-step calculations or inferential reasoning that requires synthesizing data across several rows or columns. This diversity in complexity ensures that users encounter a mix of questions, from basic checks to challenging analytical tasks, enhancing skill development. \\ 
    2.Length: Questions should be between 20 and 50 words in length. Ensure that the questions are as concise as possible while still containing the necessary details. \\
    3.Diversity: The subtypes covered by the question should contain Multi-hop Fact Checking, Value-Matching, Inference-based Fact-Checking. \\
    4.Real-World Relevance: The questions asked should match real-world scenarios, making them more practical and relatable. \\
    5.Writing-Style: Use straightforward and accessible language. Keep terminology consistent, especially for statistical or financial terms, to avoid confusion. Ensure questions are free from ambiguity and clearly specify what is expected in the answer. \\
    6.Answer Control: Ensure that the final answer format of the question can be expressed as "Final Answer: AnswerName1, AnswerName2..." form, no other form. Ensure the "AnswerName" is a number or entity name, as short as possible, without any explanation. \\

     \hline

     \# \textit{Output Control}

     Please generate it in the following format:  \\
    {[Question1]}: ... ,   \\
    {[subtype1]}: ... .   \\
    {[Question2]}: ... ,   \\
    {[subtype2]}: ... .   \\
    The subtype of each question should be one of Multi-hop Fact Checking, Value-Matching, Inference-based Fact-Checking.\\
     \hline
    \end{tabular}
\end{table*}

\begin{table*}[!ht]
    \centering
    \caption{The complete question prompt of \textbf{Numerical Reasoning}. Fill the corresponding content into the '...' positions after '[]' field.}
    \label{tab:prompt_q_numerical_reasoning}
    \begin{tabular}{m{15cm}}
    
    \hline
     \textbf{Question Generation Prompt for Numerical Reasoning} \\
    \midrule
     \# \textit{Role play} \\
     Suppose you are an expert in question annotation and your task is to generate high-quality and diverse questions based on a given task description and tabular data. \\

     \hline
     
     \# \textit{Task Descriptions} \\
     Numerical reasoning tasks in tabular data are tasks that involve analyzing, calculating, or making logical inferences based on numerical information from tabular-structured data. The task typically involves recognizing patterns, making numerical comparisons, identifying trends, predicting outcomes, or solving specific problems from large amounts of numerical data.\\

     \hline

     \# \textit{Generation Restrictions} \\
     To generate 5 questions based on the given task description and tabular data, give due consideration to the following aspects: \\
    1.Complexity: Include a range of problem complexities. Simple problems may require direct calculations or basic comparisons, while complex ones could involve multi-step calculations or inferential reasoning across multiple rows/columns.\\
    2.Length: Questions should be between 20 and 50 words in length. Ensure that the questions are as concise as possible while still containing the necessary details. \\
    3.Diversity: The subtypes covered by the question should contain Multi-hop Numerical Reasoing, Counting, Ranking, Comparison, Calculation (Numerical Calculation and Time-based Calculation). \\
    4.Real-World Relevance: The questions asked should match real-world scenarios, making them more practical and relatable. \\
    5.Writing-Style: Use straightforward and accessible language. Keep terminology consistent, especially for statistical or financial terms, to avoid confusion. Ensure questions are free from ambiguity and clearly specify what is expected in the answer. \\
    6.Answer Control: Ensure that the final answer format of the question can be expressed as "Final Answer: AnswerName1, AnswerName2..." form, no other form. Ensure the "AnswerName" is a number or entity name, as short as possible, without any explanation. \\

     \hline

     \# \textit{Output Control}

     Please generate it in the following format:  \\
    {[Question1]}: ... ,   \\
    {[subtype1]}: ... .   \\
    {[Question2]}: ... ,   \\
    {[subtype2]}: ... .   \\
    The subtype of each question should be one of Multi-hop Numerical Reasoning, Counting, Ranking, Comparison, Calculation.\\
     \hline
    \end{tabular}
\end{table*}

\begin{table*}[h]
    \centering
    \caption{The complete question prompt of \textbf{Data Analysis}. Fill the corresponding content into the '...' positions after '[]' field.}
    \label{tab:prompt_q_data_analysis}
    \begin{tabular}{m{15cm}}
    
    \hline
     \textbf{Question Generation Prompt for Data Analysis} \\
    \midrule
     \# \textit{Role play} \\
     Suppose you are an expert in question annotation and your task is to generate high-quality and diverse questions based on a given task description and tabular data. \\

     \hline
     
     \# \textit{Task Descriptions} \\
     The task of Data Analysis in tabular data is the process of systematically examining and transforming structured data to extract meaningful information, discover patterns in the data, support decision making, or test hypotheses. It aims to mine key features or trends from data, reveal hidden relationships or anomalies, predict future changes, and provide solutions to problems through quantitative or qualitative methods. This process typically outputs results in a clear and actionable form, providing a solid foundation for deeper understanding of the data.\\

     \hline

     \# \textit{Generation Restrictions} \\
     To generate 5 questions based on the given task description and tabular data, give due consideration to the following aspects: \\
    1.Complexity: Include a range of problem complexities to suit different levels of data analysis skill. Simple tasks should involve straightforward operations, such as calculating averages, identifying basic trends (e.g., increasing or decreasing patterns), or performing direct comparisons between values in a single column. Intermediate tasks might require analyzing relationships between variables using correlation or group-level aggregations, identifying potential impacts, or summarizing data patterns across multiple rows or columns. Complex tasks could involve multi-step processes such as performing causal analysis, anomaly detection requiring cross-referencing multiple datasets, or building predictive models that integrate trends and relationships.  \\
    2.Length: Questions should be between 20 and 50 words in length. Ensure that the questions are as concise as possible while still containing the necessary details. \\
    3.Diversity: The subtypes covered by the question should contain Rudimentary Analysis, Summary Analysis, Predictive Analysis, Exploratory Analysis, Anomaly Analysis. \\
    4.Real-World Relevance: The questions asked should match real-world scenarios, making them more practical and relatable. \\
    5.Writing-Style: Use straightforward and accessible language. Keep terminology consistent, especially for statistical or financial terms, to avoid confusion. Ensure questions are free from ambiguity and clearly specify what is expected in the answer.  \\
    6.Question example: Rudimentary Analysis: "What is the mean and standard deviation of the Year built column?", "Which state or region has the highest proportion of Military MPs to total MPs, and what is the percentage?". Summary analysis: "Can you provide a descriptive explanation of the table, including the main columns and some basic insights?", "Can you provide a detailed description of the table, including explanations for each main column and highlight any notable trends or insights from the data?". Predictive Analysis: "Based on the historical population growth from 1956 to 2006, what could be the projected population of Tabriz in 2026?". Exploratory Analysis: "How does the number of examinees affect the pass percentage over the years?", "Does a higher crude birth rate causally influence the natural change in population?". Anomaly Analysis: "What are the anomalies in the viewership data for the TV episodes?", "Can you identify which surname data points deviate significantly from the norm?". \\

     \hline

     \# \textit{Output Control}

     Please generate it in the following format:  \\
    {[Question1]}: ... ,         {[subtype1]}: ... . \\
    {[Question2]}: ... ,         {[subtype2]}: ... .   \\
    The subtype of each question should be one of Rudimentary Analysis, Summary Analysis, Predictive Analysis, Exploratory Analysis, Anomaly Analysis.\\
     \hline
    \end{tabular}
\end{table*}

\begin{table*}[!ht]
    \centering
    \caption{The complete question prompt of \textbf{Chart Generation}. Fill the corresponding content into the '...' positions after '[]' field.}
    \label{tab:prompt_q_chart_generation}
    \begin{tabular}{m{15cm}}
    
    \hline
     \textbf{Question Generation Prompt for Chart Generation} \\
    \midrule
     \# \textit{Role play} \\
     Suppose you are an expert in question annotation and your task is to generate high-quality and diverse questions based on a given task description and tabular data. \\

     \hline
     
     \# \textit{Task Descriptions} \\
     The task of chart generation in tables is to help users quickly understand and analyze data by converting tabular data into charts or other graphical representations that visualize relationships, trends, or characteristics of the data. The goal is to simplify complex data patterns so that distributions, trends, comparative differences, or correlations between variables can be more easily perceived. Typically, this type of task requires clarifying the analysis objectives, choosing the appropriate chart type (e.g., bar charts, line charts, pie charts, or scatter plots, etc.), and cleaning and processing the data to ensure that the final visualization results are clear, accurate, and able to convey information effectively. This approach not only enhances the intuition and efficiency of data presentation, but also facilitates further analysis and decision-making.\\

     \hline

     \# \textit{Generation Restrictions} \\
     To generate 5 questions based on the given task description and tabular data, give due consideration to the following aspects: \\
    1.Complexity: Include a range of problem complexities. Simple problems may only require extracting data and drawing a picture, while complex problems require understanding the data items that need to be used in the problem and processing them before drawing the picture. \\
    2.Length: Questions should be between 20 and 50 words in length. Ensure that the questions are as concise as possible while still containing the necessary details. \\
    3.Diversity: The subtypes covered by the question should contain LineChart Generation, BarChart Generation, ScatterChart Generation, PieChart Generation. \\
    4.Real-World Relevance: The questions asked should match real-world scenarios, making them more practical and relatable. \\
    5.Writing-Style: Use straightforward and accessible language. Keep terminology consistent, especially for statistical or financial terms, to avoid confusion. Ensure questions are free from ambiguity and clearly specify what is expected in the answer. It needs to be made clear in the question what type of chart is being plotted, such as "According to the table, draw a bar chart to illustrate ..." and "Please help me draw a line chart showing ...".  \\
    6.Answer Control: Ensure the final answer format is the python code block that can generate the chart correctly. \\

     \hline

     \# \textit{Output Control}

     Please generate it in the following format:  \\
    {[Question1]}: ... ,   \\
    {[subtype1]}: ... .   \\
    {[Question2]}: ... ,   \\
    {[subtype2]}: ... .   \\
    The subtype of each question should be one of LineChart Generation, BarChart Generation, ScatterChart Generation, PieChart Generation.\\
     \hline
    \end{tabular}
\end{table*}

\begin{table*}[!ht]
    \centering
    \caption{The complete answer prompt of \textbf{Fact Checking}. Fill the corresponding content into the '...' positions after '[]' field.}
    \label{tab:prompt_a_fact_checking}
    \begin{tabular}{m{15cm}}
    
    \hline
     \textbf{Answer Generation Prompt for Fact Checking} \\
    \midrule
     \# \textit{Role play} \\
     Suppose you are an expert in table analysis and your task is to provide answers to questions based on the content of the table. \\

     \hline

     \# \textit{Chain-of-Thought} \\
     Let’s think step by step as follows and make the most of your strengths as a table analysis expert: \\
    1.Fully understand the question and extract the necessary information from it. \\
    2.Clearly and comprehensively understanding the content of  the table, including the structure of the table, the meaning and formatting of each row and column header(Note: There is usually summative cell in the table, such as all, combine, total, sum, average, mean, etc. Please pay careful attention to the flag information in the row header and column header, this information can help you to skip many operations.) \\
    3.Based on the question, select the row and column headers in the table that are most relevant to it and find the corresponding cells based on them. \\
    4.According to the requirements of the question, perform statistical, calculation, ranking, or other operations on the cells you selected, and output of the answer in the format specified by the definition. \\

     \hline

     \# \textit{Output Control} \\
    1.First, you need to output your reasoning steps according to the question and table itself. The reasoning steps should follow the format below: [Reasoning steps for this question are as following: 1.First, we need to... 2.We need to...]. output steps until final answers get solved. \\
    2.Then, you need to output the final answer. The final answer should follow the format below: [Answer Format] Final Answer: AnswerName1, AnswerName2... Ensure the final answer format is the last output line and can only be in the "Final Answer: AnswerName1, AnswerName2..." form, no other form.  \\
    3.Ensure the "AnswerName" is a number or entity name, as short as possible, without any explanation. Give the final answer to the question directly without any explanation. If the question is judgmental, please answer 'Yes' or 'No'.
    Let's get start! \\
    {[Question]}: ... \\
     \hline
    \end{tabular}
\end{table*}

\begin{table*}[!ht]
    \centering
    \caption{The complete answer prompt of \textbf{Numerical Reasoning}. Fill the corresponding content into the '...' positions after '[]' field.}
    \label{tab:prompt_a_numerical_reasoning}
    \begin{tabular}{m{15cm}}
    
    \hline
     \textbf{Answer Generation Prompt for Numerical Reasoning} \\
    \midrule
     \# \textit{Role play} \\
     Suppose you are an expert in table analysis and your task is to provide answers to questions based on the content of the table. \\

     \hline

     \# \textit{Chain-of-Thought} \\
     Let’s think step by step as follows and make the most of your strengths as a table analysis expert: \\
    1.Fully understand the question and extract the necessary information from it. \\
    2.Clearly and comprehensively understanding the content of  the table, including the structure of the table, the meaning and formatting of each row and column header(Note: There is usually summative cell in the table, such as all, combine, total, sum, average, mean, etc. Please pay careful attention to the flag information in the row header and column header, this information can help you to skip many operations.) \\
    3.Based on the question, select the row and column headers in the table that are most relevant to it and find the corresponding cells based on them. \\
    4.According to the requirements of the question, perform statistical, calculation, ranking, or other operations on the cells you selected, and output of the answer in the format specified by the definition. \\

     \hline

     \# \textit{Output Control} \\
    1.First, you need to output your reasoning steps according to the question and table itself. The reasoning steps should follow the format below: [Reasoning steps for this question are as following: 1.First, we need to... 2.We need to...]. output steps until final answers get solved. \\
    2.Then, you need to output the final answer. The final answer should follow the format below: [Answer Format] Final Answer: AnswerName1, AnswerName2... Ensure the final answer format is the last output line and can only be in the "Final Answer: AnswerName1, AnswerName2..." form, no other form.  \\
    3.Ensure the "AnswerName" is a number or entity name, as short as possible, without any explanation. Give the final answer to the question directly without any explanation. Note: If the final answer has multiple decimals, retain two decimals.  \\
    Let's get start! \\
    {[Question]}: ... \\
     \hline
    \end{tabular}
\end{table*}

\begin{table*}[!ht]
    \centering
    \caption{The complete answer prompt of \textbf{Data Analysis}. Fill the corresponding content into the '...' positions after '[]' field.}
    \label{tab:prompt_a_data_analysis}
    \begin{tabular}{m{15cm}}
    
    \hline
     \textbf{Answer Generation Prompt for Data Analysis} \\
    \midrule
     \# \textit{Role play} \\
     Suppose you are an expert in table analysis and your task is to provide answers to questions based on the content of the table. \\

     \hline

     \# \textit{Chain-of-Thought} \\
     Let’s think step by step as follows and make the most of your strengths as a table analysis expert: \\
    1.Fully understand the question and extract the necessary information from it. \\
    2.Clearly and comprehensively understanding the content of  the table, including the structure of the table, the meaning and formatting of each row and column header(Note: There is usually summative cell in the table, such as all, combine, total, sum, average, mean, etc. Please pay careful attention to the flag information in the row header and column header, this information can help you to skip many operations.) \\
    3.Based on the question, select the row and column headers in the table that are most relevant to it and find the corresponding cells based on them. \\
    4.According to the requirements of the question, perform statistical, calculation, ranking, or other operations on the cells you selected, and output of the answer in the format specified by the definition. \\

     \hline

     \# \textit{Output Control} \\
    1.First, you need to output your reasoning steps according to the question and table itself. The reasoning steps should follow the format below: [Reasoning steps for this question are as following: 1.First, we need to... 2.We need to...]. output steps until final answers get solved. \\
    2.Then, you need to output the final answer. The final answer should follow the format below: [Answer Format] Final Answer: AnswerName1, AnswerName2... Ensure the final answer format is the last output line and can only be in the "Final Answer: AnswerName1, AnswerName2..." form, no other form.  \\
    3.The "AnswerName" should represent the primary result of the rudimentary analysis, such as a number or an entity name, expressed as concisely as possible. Provide the final answer directly without additional explanation or extra output. \\
    Let's get start! \\
    {[Question]}: ... \\
     \hline
    \end{tabular}
\end{table*}

\begin{table*}[!ht]
    \centering
    \caption{The complete answer prompt of \textbf{Chart Generation}. Fill the corresponding content into the '...' positions after '[]' field.}
    \label{tab:prompt_a_chart_generation}
    \begin{tabular}{m{15cm}}
    
    \hline
     \textbf{Answer Generation Prompt for Chart Generation} \\
    \midrule
     \# \textit{Role play} \\
     Suppose you are an expert in table analysis and your task is to provide answers to questions based on the content of the table. \\

     \hline

     \# \textit{Chain-of-Thought} \\
     Let’s think step by step as follows and make the most of your strengths as a table analysis expert: \\
    1.Fully understand the question and extract the necessary information from it. \\
    2.Clearly and comprehensively understanding the content of  the table, including the structure of the table, the meaning and formatting of each row and column header(Note: There is usually summative cell in the table, such as all, combine, total, sum, average, mean, etc. Please pay careful attention to the flag information in the row header and column header, this information can help you to skip many operations.) \\
    3.Based on the question, select the row and column headers in the table that are most relevant to it and find the corresponding cells based on them. \\
    4.According to the requirements of the question, perform statistical, calculation, ranking, or other operations on the cells you selected, and output of the answer in the format specified by the definition. \\

     \hline

     \# \textit{Output Control} \\
    1.First, you need to output your [reasoning steps] according to the question and table itself. The reasoning steps should follow the format below: [Reasoning steps for this question are as following: 1.First, we need to... 2.We need to...]. output steps until final answers get solved. Then, you need to output the final answer.  \\
    2.The final answer should follow the format below and ensure the first three code lines is exactly the same with the following code block: {[Answer Format]} python import pandas as pd import matplotlib.pyplot as plt df = pd.read\_excel('table.xlsx') ... plt.show(). Ensure the code can generate the chart correctly and output this code block completely.  \\
    3.You should take the values needed to draw the chart directly from the table and write them into the code block,  e.g. name1: value1, name2: value2...  \\
    4.Then transform it into a string with newlines represented as '\string\n', indents represented as '\string\t' and no comments. Ensure code block and corresponding string are right. Do NOT output {[answer format]}.  \\
    5.Ensure that the X-axis used for drawing in the code is arranged in ascending alphabetical or numerical order. Ensure the last line in python code can only be "plt.show()", no other from. Give the final answer to the question directly without any explanation. \\
    Let's get start! \\
    {[Question]}: ... \\
     \hline
    \end{tabular}
\end{table*}

\begin{table*}[!ht]
    \centering
    \caption{The \textbf{First-Round Tree-Based} answer prompt. Fill the corresponding content into the '...' positions after '[]' field.}
    \label{tab:first_round_tree}
    \begin{tabular}{m{15cm}}
    
    \hline
     \textbf{First-Round Tree-Based Answer Prompt} \\
    \midrule
     \# \textit{Role play} \\
     You are tasked with performing detailed table analysis. Your task is to generate a hierarchical tree structure for the top-row and left-column headers based on a LaTeX syntax complex table. \\

     \hline

     \# \textit{Task Description} \\
     {[Reasoning Steps]} \\
    Your thought process is as follows: \\
    1.Understand the Table Structure: Provide a comprehensive description of the table, including the various levels of row and column headers and their corresponding meanings. Construct two distinct hierarchical trees: one for the row headers and one for the column headers. Each tree should accurately represent the levels and relationships of the headers. \\
    2.Traverse the Table: Analyze each row and column header to extract its content, indentation, and positions in the table. Identify merged cells and indentation, as they often indicate hierarchical relationships. Determine the parent-child relationships based on these visual cues and arrange the data under the correct parent node in both row and column header trees. \\
    3.Validate the Hierarchical Relationships: Iterate through both the row header tree and column header tree. Verify that the parent-child relationships are accurate and that the nodes are correctly placed within their respective hierarchies. \\

     \hline

     \# \textit{Node Definition} \\
    You will be provided with a table in LaTeX format. The table may contain complex structures, such as merged or nested cells. Your task is to encode each node of table header as a tuple T(t1, t2, t3, t4). \\
    The first element t1 indicates it represents row header (R) or column header (C), along with its corresponding level. \\ 
    The second element t2 and third element t3 represent its start and end positions, while the fourth element t4 contains the value from the table. For example, a tuple (R0, 1, 2, City) indicates that it is a row header (R) at level 0, spanning from row 1 to row 2, with the value City. \\
    Please Convert the table headers to list L={[T1, T2, ...]}. \\
     \hline

    \# \textit{Tree Generate} \\
    1. Divide the tuples list L into groups based on their levels, such that all tuples with the same level are grouped together. Add a special ROOT node for rows and columns, each with a level of "-1". \\
    2. For each tuple A in L. If the start and end positions of A are equal, mark A as a leaf node. \\
    3. Otherwise, compare its T2 and T3 values with every closest higher-level and same flag tuple B. If tuple A is within the range of tuple B, then B is the parent-header of A. \\
    4. Repeat steps 2 and 3 iteratively until all tuples in L are linked to their respective parent nodes (Tuples without parent node are linked to the ROOT node), forming a hierarchical Table-Header Tree H. \\

     \hline

    \# \textit{Output Control} \\
    Next, we will provide a table for you to analyze the hierarchical structure for the table and please organize the table header tuples as a tree, which can help you better understand the table structure. \\
    You should clearly and comprehensively understand the content of the table, including the structure of the table, the meaning and formatting of each row and column header (Note: There is usually summative cell in the table, such as all, combine, total, sum, average, mean, etc. Please pay careful attention to the flag information in the row header and column header, this information can help you to skip many operations.) \\
    
    Please check the constructed tree structure carefully and make sure that you have not missed any information in the contents of the table.\\
    
    Let's get started! \\
    {[TABLE]}: ... \\

     \hline
    \end{tabular}
\end{table*}

\begin{table*}[!ht]
    \centering
    \small
    \caption{The \textbf{Second-Round Tree-Based} answer prompt. Fill the corresponding content into the '...' positions after '[]' field.}
    \label{tab:second_round_tree}
    \begin{tabular}{m{15cm}}
    
    \hline
     \textbf{Second-Round Tree-Based Answer Prompt} \\
    \midrule
     \# \textit{Role play} \\
     You are a table analyst. Your task is to first extract relevant keywords based on the questions posed, identify related content from the previous tables, and match it with the corresponding headers in the structure tree. Then you need to answer questions based on the provided table content. \\

     \hline

     \# \textit{Thinking Guidelines and Output Format Control} \\
     1.Understand the Question: Begin by carefully reading the question to extract the essential information needed for answering. This helps ensure that you focus on the right aspects of the table in the next steps. \\
    2.Analyze the Table Content: Thoroughly examine the structure tree and original content of the table, paying close attention to both row and column headers, which may include special indicators such as "Total," "Sum," "Average," or other summary metrics. Be mindful of any rows or columns dedicated to aggregates, as these can provide quick answers without the need for detailed calculations. It's also crucial to recognize that the table might have complex structures, such as merged cells or semantic nesting, which could influence the interpretation of the data. \\
    3.Identify Relevant Data: With a clear understanding of the question, identify the rows and columns in the table that are most relevant to the inquiry. This involves locating the cells that correspond to the relevant headers, ensuring the selected data is directly related to the question at hand. \\
    4.Perform Necessary Analysis or Calculations: Once the relevant data is identified, perform any required operations, such as statistical analysis, mathematical calculations, ranking, or other necessary procedures. This will help you derive the needed insights and provide a comprehensive answer. \\

     \hline

     \# \textit{Output Action Pattern} \\
    Your output should follow a React-like pattern of thinking, which includes one or more cycles of "Thought/Action/Result", ultimately leading to a "Final Answer" on the last line. \\
    {[Action Patterns]} \\
    1.Thought: Consider the next action based on the result of the previous one. \\
    2.Action: The action should always be a single processing action. \\
    3.Result: Simulate the action result, analyze the result, and decide whether to continue or stop. \\
    (This "Thought/Action/Result" cycle can repeat multiple times.) \\
    Verify the table, observations, and question thoroughly before providing the final answer. \\

    \hline

    \# \textit{Output Format} \\
    When answering, if the final answer comes from the original format in the table, please use the original format from the table without modifying it. \\
    Below is an example of an output format. You need to first output the relevant keywords, headers, and content related with the question, then go through a multi-round interaction of thought/action/result, and finally provide the final answer. \\

    \hline

    \# \textit{Output Example} \\
    Relevant Keywords: Keywords related to the table in question. \\
    Relavant Table Headers: column/row headers related with the question. \\
    Relavant Content: related table content. \\
    Thought: Your first round of thinking. \\
    Action: The action of your first round. \\
    Result: The observation and result of your first round of simulation. \\
    Thought: Your second round of thinking. The 'Thought/Action/Result' cycle can repeat 1 or more times until the final answer is reached. \\
    Action: The action of your second round. \\
    Result: The observation and result of your second round of simulation. \\
    Final Answer: Your output result, following the format "Final Answer: AnswerName1, AnswerName2...". The "AnswerName" should be a number or entity name, as short as possible. \\
    Let's get start! \\
    {[Question]}: ... \\

     \hline
    
    \end{tabular}
\end{table*}

\begin{table*}[!ht]
    \centering
    \caption{The representative GPT-Eval prompt of \textbf{Data Analysis}. Fill the corresponding content into the '...' positions after '[]' field.}
    \label{tab:GPT-Eval_prompt}
    \begin{tabular}{m{15cm}}
    
    \hline
     \textbf{Representative GPT-Eval Prompt for Data Analysis} \\
    \midrule
      Suppose you are an expert in table analysis and your task is to rate the user answer on one metric based on the table content, question and corresponding reference answer.   \\
    You will be given table content and a question about rudimentary analysis of the table. And the corresponding reference answer to a question and the answer from the user.  \\
    Your task is to rate the answer on one metric.  \\
    Please make sure you read and understand these instructions carefully. Please keep this document open while reviewing, and refer to it as needed.  \\
    Evaluation Criteria:  \\
    Correctness (1-100) - the answer should be as close as possible to the reference answer, with perfectly equal answers receiving full marks, smaller differences receiving higher marks, and larger differences receiving only lower marks. \\
    Evaluation Steps:  \\
    1. Read the table carefully and fully understand the contents of the table. \\
    2. Read the result and compare it to the reference answer and the table. Determine if the answer is correct and if not score it based on how different it is from the correct answer. \\
    3. Assign a score for correctness on a scale of 0 to 100, where 0 is the lowest and 100 is the highest based on the Evaluation Criteria.  \\  
    {[Question]}: ... , \\
    {[Reference Answer]}: ... , \\
    {[User Answer]}: ... . \\
    Emphasize: you need to make sure your final answer is formatted in this way: [Score]: xx/100\\
     \hline
    \end{tabular}
\end{table*}

\begin{figure*}
    \centering
    \includegraphics[width=1\linewidth]{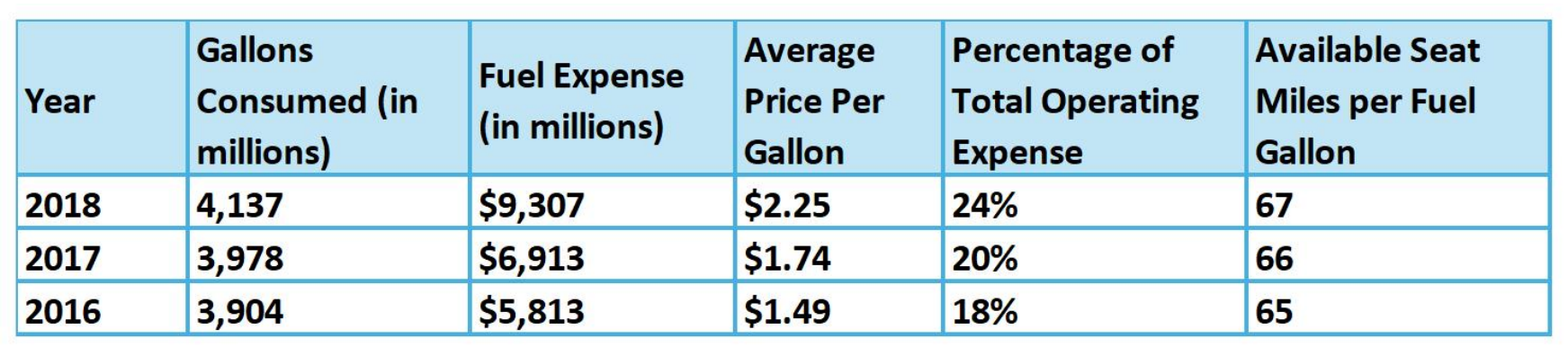}
    \caption{The flat table sample in TableBench~\citep{tablebench}.}
    \label{fig:tableBench_table_sample}
\end{figure*}

\begin{figure*}
    \centering
    \includegraphics[width=1\linewidth]{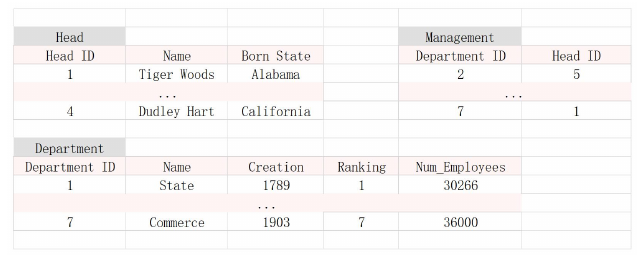}
    \caption{The multiple table sample in MMQA~\citep{MMQA}.}
    \label{fig:MMQA_multi_table_sample}
\end{figure*}

\end{document}